\PassOptionsToPackage{capitalize,noabbrev}{cleveref}
\documentclass[]{fairmeta}

\usepackage{microtype}
\usepackage{graphicx}
\usepackage{subcaption}
\usepackage{booktabs} 
\usepackage{cell}
\usepackage[T1]{fontenc}
\usepackage{amsmath}
\usepackage{amssymb}
\usepackage{amsthm}
\usepackage{mathtools}
\usepackage[most]{tcolorbox}
\usepackage{multirow}
\usepackage{enumitem}
\usepackage{wrapfig}
\usepackage{makecell} 

\usepackage{hyperref}

\theoremstyle{plain}

\theoremstyle{definition}

\theoremstyle{remark}

\usepackage[textsize=tiny]{todonotes}

\title{Asking the Right Questions: Improving\\Reasoning with Generated Stepping Stones}

\author[1,2,*,\dagger]{Hengyuan Hu}
\author[1,3,*]{Tingchen Fu}
\author[1,\dagger]{Minqi Jiang}
\author[1,]{Alexander H Miller}
\author[1]{Yoram Bachrach}
\author[1,3]{Jakob Nicolaus Foerster}

\affiliation[1]{FAIR at Meta}
\affiliation[2]{Stanford University}
\affiliation[3]{University of Oxford}
\contribution[*]{Joint first author}
\contribution[\dagger]{Work done while at Meta}

\abstract{
Recent years have witnessed tremendous progress in enabling LLMs to solve complex reasoning tasks such as math and coding. 
As we start to apply LLMs to harder tasks that they may not be able to solve in one shot, it is worth paying attention to their ability to construct intermediate stepping stones that prepare them to better solve the tasks. Examples of stepping stones include simplifications, alternative framings, or subproblems.
We study properties and benefits of stepping stones in the context of modern reasoning LLMs via ARQ (\textbf{A}sking the \textbf{R}ight \textbf{Q}uestions), a simple framework that introduces a question generator to the default reasoning pipeline. 
We first show that good stepping stone questions exist and are transferrable, meaning that good questions can be generated, and they substantially help LLMs of various capabilities in solving the target tasks. We next frame stepping stone generation as a post-training task and show that we can fine-tune LLMs to generate more useful stepping stones by SFT and RL on synthetic data.
}

\date{\today}
\correspondence{Hengyuan Hu at \email{hengyuan@stanford.edu}}

\newif\ifmeta
\metatrue

\begin{document}

\maketitle

\section{Introduction}

\ifmeta
\begin{wrapfigure}[24]{r}{0.5\textwidth}
\centering
\vspace{-6mm}
\includegraphics[width=1.0\linewidth]{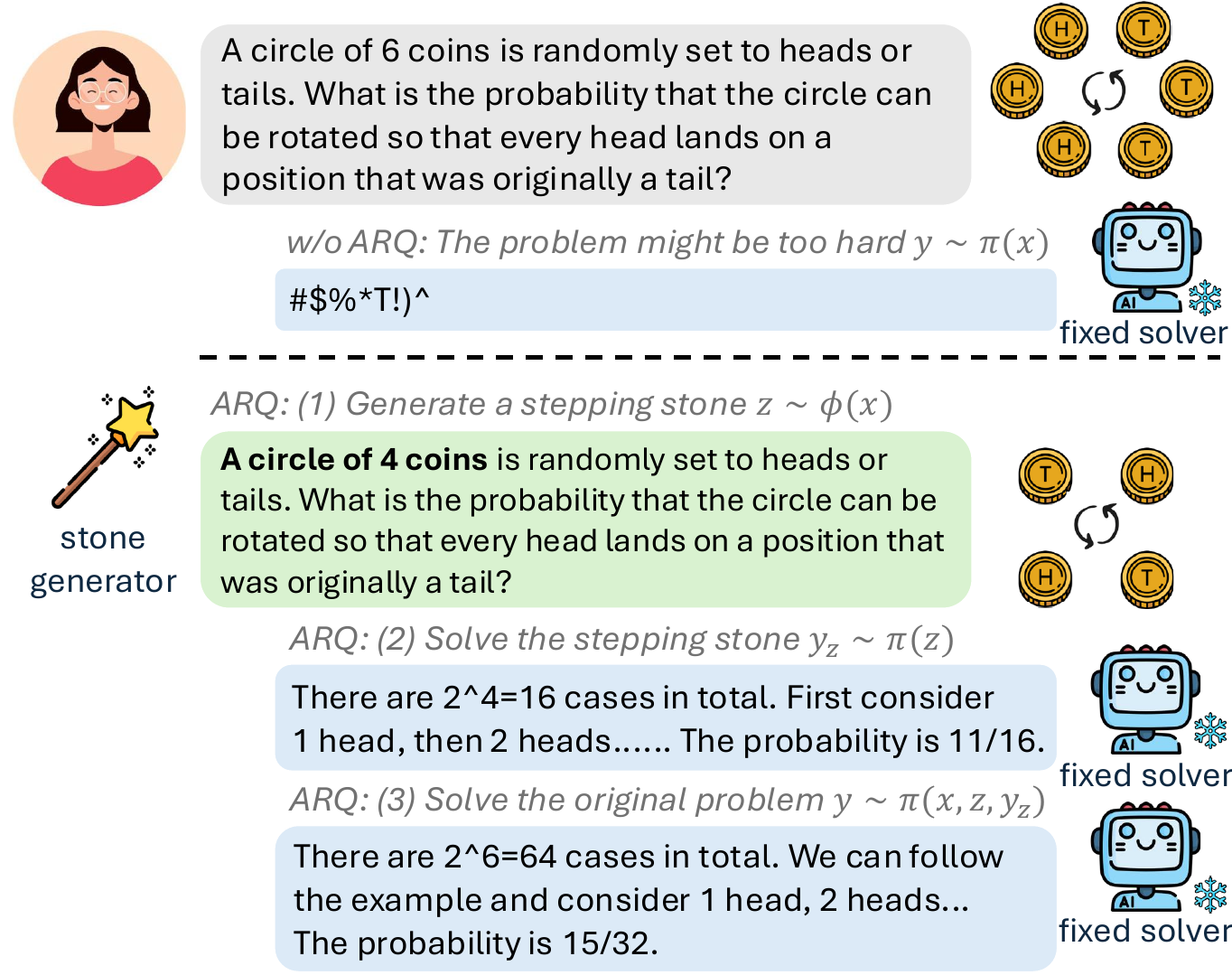}
\caption{\small Illustration of ARQ. Instead of directly solving a task that may be too hard for a given solver (top), ARQ (bottom) adds a question-asking step that prompts or trains LLMs to generate \textit{stepping stone questions} which, once solved, provide guidance or inspiration for the original problem. The LLMs used to generate stepping stone questions are referred to as \textit{stepping stone generators}. In this example, the stepping stone (generated by an LLM) focuses on a special case of the original problem.}
\label{fig:method}
\end{wrapfigure}
\fi

There has been rapid progress in training large language models (LLMs) to solve increasingly complex reasoning tasks such as mathematics~\citep{openai2024openaio1card, lightman2023lets, shao2024deepseekmathpushinglimitsmathematical}, coding~\citep{chen2021codex, gehring2025rlefgroundingcodellms}, and decision-making in long-horizon agentic environments~\citep{chan2025mlebenchevaluatingmachinelearning, jimenez2024swebenchlanguagemodelsresolve}.

A common recipe for building reasoning models combines post-training with test-time scaffolding.
The first (optional) step in post-training is to curate datasets that explicitly exhibit the desired reasoning behaviors and use them for supervised finetuning (SFT). For instance, problem-solving datasets annotated with high-quality chain-of-thought (CoT) rationales~\citep{cot} can encourage pretrained LLMs to reliably express these behaviors after SFT~\citep{ye2025limo, muennighoff2025s1simpletesttimescaling}.
Then reinforcement learning (RL) can be used to reinforce effective strategies that achieve high reward~\citep{r1}.
After post-training, models can further be augmented with test-time scaffolds that provide explicit procedures for verification~\citep{zheng2023judging, huang2025winninggoldimo2025}, feedback~\citep{lee2025feedbackdescentopenendedtext}, or iterative refinement~\citep{self-refine, shinn2023reflexion}. These scaffolds further improve performance when the underlying model is capable of leveraging such behaviors~\citep{kim2025astro, shen2025satori, qin2025backtrack}.



However, most post-training and inference-time approaches have focused on reasoning behaviors that help models solve the given problem or verify proposed solutions. In contrast, the ability to ask questions --- especially to pose relevant, insight-generating questions that expand one’s knowledge and thus enable progress on problems that are otherwise too hard --- has received less attention. We believe this is due to the paradigm shift to encapsulate all reasoning behaviors into a single, long 
thinking trace. Yet, asking the right question is a crucial component of human intelligence~\citep{newell1972human}. For example, when tackling challenging tasks such as scientific research, we often construct related but more approachable \textit{stepping stone problems} to build intuition~\citep{ho2022people}.
Crucially, stepping stone problems are often not only easier to solve, but also cheaper and easier to verify, enabling faster iteration. For example, in deep learning research, practitioners routinely prototype on small datasets or simplified toy settings to validate the feasibility and correctness of ideas, assumptions, or implementations before scaling up to the target, large-scale domain~\citep{kaplan2020scaling,henighan2020scaling}.

\ifmeta\else
\begin{wrapfigure}[24]{r}{0.5\textwidth}
\centering
\vspace{-6mm}
\includegraphics[width=1.0\linewidth]{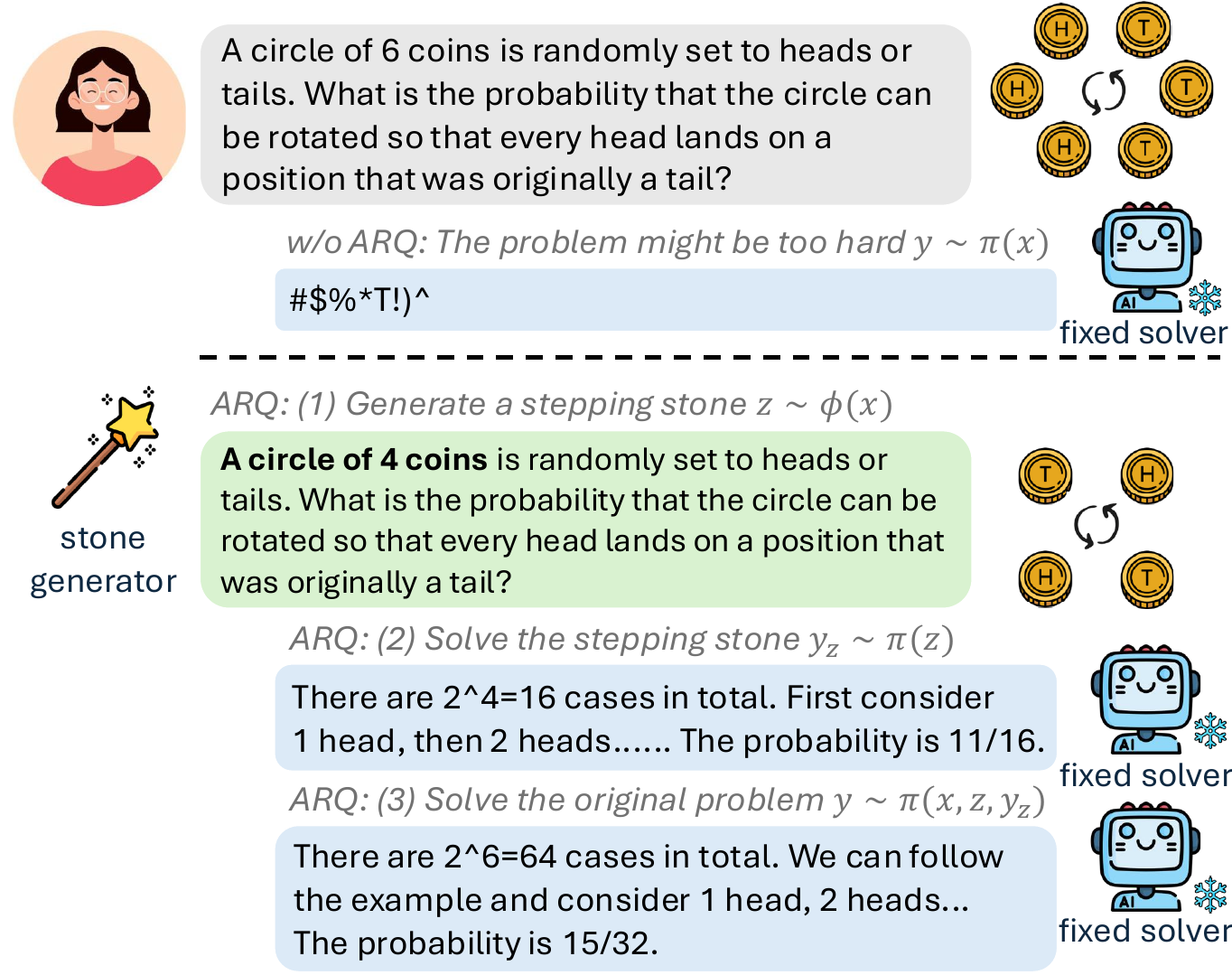}
\caption{\small Illustration of ARQ. Instead of directly solving a task that may be too hard for a given solver (top), ARQ (bottom) adds a question-asking step that prompts or trains LLMs to generate \textit{stepping stone questions} which, once solved, provide guidance or inspiration for the original problem. The LLMs used to generate stepping stone questions are referred to as \textit{stepping stone generators}. In this example, the stepping stone (generated by an LLM) focuses on a special case of the original problem.}
\label{fig:method}
\end{wrapfigure}
\fi

As LLMs are applied to longer-term reasoning scenarios~\citep{metr}, it is important to understand whether they, like humans, can also generate a useful test-time curriculum by asking the right stepping stone questions.
In this work, we propose a straightforward framework, ARQ (\textbf{A}sking the \textbf{R}ight \textbf{Q}uestions), to study this question from both inference and post-training perspectives. 
As shown in \cref{fig:method}, ARQ consists of two LLM-based modules with distinct roles: a \textit{question generator} $\phi$ and a \textit{problem solver} $\pi$. 
Given a target problem $x$, instead of directly producing a solution $y \sim \pi(x)$, ARQ first generates a stepping stone problem $z \sim \phi(x)$. 
It then samples a solution $y_{z} \sim \pi(z)$ to the stepping stone and finally solves the target problem with $(z, y_{z})$ in the context, i.e., $y \sim \pi(x; z, y_{z})$. 
For simplicity, we do not use  more advanced techniques for the LLM-based solver $\pi$, such as majority voting, since they are complementarity to our method.
We instead focus only on the question generator $\phi$ to study whether asking the right questions \textit{in isolation} can improve a given solver’s success rate on challenging tasks.

We first investigate ARQ as an inference-time scaffold on top of existing reasoning LLMs on three challenging math benchmarks, showing that good stepping stones exist and that the \textit{best stones} greatly improve the solver’s success rate by $13\%$ on average (\cref{sec:good-stone}).
We then demonstrate that these improvements are \textit{transferable}: good stepping stones are broadly helpful across multiple solvers with various levels of performance, and the gains are not due to overfitting to a specific solver~(\cref{sec:transfer}).
With this evidence, we establish stepping stone generation as a meaningful post-training task. Here, we score a given stepping stone using the expected reward of the solver when solving the target problem conditioned on this stone. We rate generations from existing LLMs with this score function, and curate a dataset for post-training. We show that additional post-training on this dataset substantially improves the ability of LLMs to ask effective stepping stone questions~(\cref{sec:post-train}). 
Finally, we extend ARQ to generate multiple stepping stones sequentially or recursively using both off-the-shelf and post-trained models (\cref{sec:more-stones}), where we find that additional stepping stones can further improve performance. 
We thus believe that ARQ offers an encouraging step into an underexplored research direction: what questions are worth asking, and what principled methods can train LLMs to generate those questions that are most beneficial for a given downstream task?

\section{Related Work}

\textbf{Post-training LLMs for Reasoning.}
The standard paradigm for enhancing the reasoning capabilities of LLMs during post-training typically consists of two stages: (1) Supervised Fine-Tuning (SFT) on curated, high-quality reasoning data to elicit CoT, and (2) Reinforcement Learning (RL) to further refine reasoning behaviors via human or verifiable feedback~\citep{li2025system,kumar2025llm}. 
In the SFT stage, a primary challenge is the acquisition of high-quality rationales. Current literature addresses this either through self-bootstrapping techniques~\citep{zelikman2022starbootstrappingreasoningreasoning, zhang2024rest} or by curating compact, high-signal datasets that prioritize diversity and quality~\citep{muennighoff2025s1simpletesttimescaling, ye2025limo, li2025llms, zhao2025million}.
Subsequently, the RL stage aligns these models with human preferences or objective reward functions. This is achieved through on-policy methods using learned reward models (RLHF) \citep{ouyang2022traininglanguagemodelsfollow}, or via Direct Preference Optimization (DPO) \citep{rafailov2023direct}, which bypasses the complexities of on-policy RL while enhancing reasoning effectiveness. Apart from human feedback, verifiable reward RL~\citep{lambert2025tulu} utilizes automated, deterministic objectives such as exact-match correctness in mathematics \citep{r1, team2025kimi} or unit-test pass rates in programming \citep{xue2025simpletir}—to provide reliable and scalable reward signals for large-scale optimization.

While prior research has largely focused on developing end-to-end solvers, we investigate a less-explored dimension of reasoning model: the ability to formulate ``stepping stone'' questions. By generating these intermediate queries, our approach incrementally builds the knowledge and skills necessary to tackle problems that exceed the capabilities of current LLMs. 

\textbf{Inference-time Scaffolds for Reasoning.}
Recent research has explored integrating LLMs into scaffolds (lightweight procedures or programs that wrap an LLM as a component.) ~\citep{welleck2024decoding, wang2025survey}, aiming to improve LLM performance by adding a layer of search or planning algorithms at inference time without modifying the underlying weights. Notable examples include parallel sampling techniques such as majority voting and Best-of-N aggregation \citep{wang2023selfconsistencyimproveschainthought, brown2024largelanguagemonkeysscaling}, iterative refinement loops \citep{self-refine, snell2024scalingllmtesttimecompute}, and tree-based search strategies \citep{yao2023treethoughtsdeliberateproblem, besta2024got}. ARQ introduces a novel question-asking step that can be integrated as an additional operator within iterative or tree-based search. Existing techniques, like majority voting, can also be leveraged to further enhance the solver of ARQ.

The inference part of our approach is closely related to prompting techniques~\citep{press2023measuringnarrowingcompositionalitygap,yasunaga2024llm-analogical-reasoner,zhou2023leasttomost} that utilize in-context learning to induce models to generate auxiliary questions before addressing the primary task. For instance, Self-Ask and similar works ~\citep{press2023measuringnarrowingcompositionalitygap, liu-etal-2022-generated,zheng2024take, zhou2024selfdiscovery} prompt LLMs to retrieve intermediate facts or knowledge (e.g., a person's age) before resolving relevant queries. Similarly, Analogical Reasoners~\citep{yasunaga2024llm-analogical-reasoner} instructs models to generate and solve a related sub-problem as a precursor to the target math problem. This can be viewed as a simpler version of the ARQ inference procedure, condensed into a single step rather than a multi-stage process. Furthermore, Least-to-Most~\citep{zhou2023leasttomost} and Decomposed Prompting~\citep{khot2023decomposed} employ a similar philosophy by decomposing tasks into a linear sequence of sub-problems. 

These existing methods are purely inference-time approaches that aim to improve performance by eliciting extended CoT traces through prompting alone.
However, later in \cref{sec:limitation-inference-only}, we observe that these prompt-based approaches are not beneficial anymore when applied to the latest reasoning LLMs that already produce long CoTs. 
ARQ can be seen as a modernization of these approaches with multi-step reasoning instead of purely prompting. 
Our analysis also shows the limitation of existing LLMs in terms of their question asking capability. 

\textbf{Synthetic Task \& Data Generation for LLMs.}
Synthetic data and task has emerged as a scalable and cost-effective paradigm for teaching LLMs specialized skills without relying on scarce expert data. A common method involves using a frontier model to synthesize both novel tasks and their solutions on the basis of existing seed tasks~\citep{xu2024wizardlm,luo2023wizardcoder,tian2024toward,li2025questa}, followed by a rigorous curation process to select high-quality training pairs for fine-tuning~\citep{wang2023selfinstructaligninglanguagemodels, chen2025self}. Within the reasoning domain specifically, recent work has expanded mathematical corpora through multi-perspective rewriting \citep{yu2024metamathbootstrapmathematicalquestions} and back-translation of questions from augmented reasoning traces \citep{lu2024mathgeniegeneratingsyntheticdata}.

In this work, we formalize stepping stone generation as a novel reasoning task: the objective is to generate the intermediate query that maximizes the probability of the solver reaching the solution for a given target question. The reward can be derived from Monte Carlo rollouts of the solver conditioned on the stepping stone question. Our fine-tuning results indicate that such a task enables models to generate more strategic stepping stone questions.

\section{Method}

When humans tackle long-term reasoning tasks such as proving novel theorems or conducting scientific research, we commonly approach the challenging problem by asking less complicated questions to decompose the task, to reduce the cost of experimentation, or to build our intuitions. These intermediate problems serve as \textit{stepping stones} toward solving the ultimate problem. This ability to generate diverse yet useful stepping stones is also important for algorithms such as search and planning~\citep{yao2023treethoughtsdeliberateproblem}.
In this work, we propose ARQ (\textbf{A}sking the \textbf{R}ight \textbf{Q}uestions), a framework that includes both an inference procedure (\cref{sec:method-inference}) and a synthetic data curation pipeline (\cref{sec:method-train}) for post-training LLMs to generate more useful stepping stones.

\subsection{The Inference Procedure of ARQ}
\label{sec:method-inference}

We formalize LLM-based problem solving with the following components. Let $x$ denote a problem stated in natural language, and $\pi$ be an LLM-based stochastic solver from which we sample solutions, $y \sim \pi(x)$. The final solution quality is evaluated by an external reward function $R(y)$.

\cref{fig:method} illustrates the core idea of ARQ.
We introduce a new LLM-based stochastic generator $\phi$ that produces stepping stone problems (``stones'') conditioned on the target problem $x$, denoted as $z \sim \phi(x)$. In the generator prompt, we define stepping stones as problems that are relevant to $x$ but easier to solve, such as special cases of $x$, or closely related problems that share the same underlying solution strategy and can be used to check whether a proposed approach is likely to work. 
Then ARQ uses the solver $\pi$ to sample a solution to this stepping stones $y_z \sim \pi(z)$. 
Finally, we prepend the generated stepping stone and its solution to the target problem in the prompt, and samples a solution from the solver, $y \sim \pi(x; z, y_z)$. 
The prompts for generator and solver are provided in~\cref{fig:stone-gen-prompt-full,fig:stone-solve-prompt,fig:final-solve-prompt} of~\cref{sec:appendix-prompt}.

Note that both the generator $\phi$ and the solver $\pi$ are both reasoning LLMs that generate CoT before producing final outputs. We remove the thinking tokens before passing the outputs to the next module to simplify the context, as we find it leads to better performance. Since we do not have a ground-truth reward function for the generated problems, we treat the sampled solutions as if they were always correct. It is possible to apply additional steps, such as iterative refinement~\citep{self-refine} or LLM-as-a-judge~\citep{zheng2023judging}, to improve the correctness of the sampled solutions. 
However, for simplicity, we focus on the question-generation component and keep the solver procedure simple.

We can further apply this procedure multiple times to generate a collection of stepping stones and build a test-time curriculum. In this paper, we will consider two strategies: generating stepping stones sequentially, or generating them recursively. 
The number of stepping stones is a predetermined hyperparameter.
In the sequential version, every time we generate a new stepping stone, the LLM takes as input both the existing ones and the target problem, $z_i \sim \phi(x; z_1, \dots, z_{i-1})$, and comes up with a new problem that aims to bridge any gap between them. The stepping stones are solved by the same solver, $y_{z_i} \sim \pi(z_i)$, and the full sequence of stones and their solutions forms the context for solving the target problem, $y \sim \pi(x; z_1, y_{z_1}, \dots, z_i, y_{z_i})$.
In the recursive version, we treat the previously generated stepping stone as the new target problem, $z_i \sim \phi(z_{i-1})$ and $z_0 = x$. The stepping stones are solved in \emph{reverse} order, with the immediately subsequent stone and its solution in the context, $y_{z_{i-1}} \sim \pi(z_{i-1}; z_i, y_{z_i})$. The motivation behind the recursive version is to keep constructing stepping stones that help solve previously generated stepping stones, with the target problem at the root.

\subsection{Limitation of Running ARQ on Existing LLMs}
\label{sec:limitation-inference-only}

\begin{table}[t]
\centering
\ifmeta\else
\small
\fi
\label{tab:beyond-aime-results}
\begin{tabular}{c|cccc|cc}
\toprule
Solver only & Analogical & Least-to-Most & Plan-and-Solve & ARQ & MajVote$@3$ & Self-Refine\\
\midrule
69.8\% & 58.6\% & 64.2\% & 60.7\% & 72.8\% & 70.0\% & 73.2\% \\
\bottomrule
\end{tabular}
\vspace{1mm}
\caption{\small Performance of running various test-time scaffold on BeyondAIME (100 questions) with GPT-OSS-120B (high reasoning effort). We run method 20 times on each problem. \textbf{Solver only} directly uses the LLM to solve the target question. The methods in the middle, including ARQ, share the similar idea of generating intermediate steps, either as analogical questions (\textbf{Analogical}), decomposition (\textbf{Least-to-Most}), or plan (\textbf{Plan-and-Solve}). On the right, we have two methods that directly improve the \textit{solver}. We run \textbf{MajVote} with 3 parallel rollouts to match the number of rollouts of ARQ and also run
\textbf{Self-Refine}, a well-established paradigm where the LLM can iteratively refine its previous solution. ARQ outperforms existing techniques focusing on introducing intermediate steps but slightly underperforms the mature Self-Refine method on existing LLMs.}
\vspace{-6mm}
\label{tab:inference}
\end{table}

The ARQ procedure described above would ideally be an inference only approach applicable to any existing LLMs. However, due to the novelty nature of this question generation step, we find that existing LLMs are not guaranteed to benefit from such an inference time procedure off-the-shelf as they generate a mixture of good and bad stepping stones. To illustrate this point, we run a pilot experiment in~\cref{tab:inference} using GPT-OSS-120B (high reasoning effort), an open-weight LLM that is more likely to ask useful questions due to its strong reasoning performance. We use BeyondAIME~\citep{seed2025seed15thinkingadvancingsuperbreasoning-beyondaime} as the test dataset because it is difficult enough that the LLM alone (\textbf{Solver only}) only achieves a moderate $69.8\%$ success rate. We compare ARQ against works that share a similar spirit of introducing intermediate components to facilitate reasoning, such as Analogical~\citep{yasunaga2024llm-analogical-reasoner}, Least-to-Most~\citep{zhou2023leasttomost}, and Plan-and-Solve~\citep{wang2023plan}, as well as methods that directly operate on the solver, such as Majority voting~\citep{wang2023selfconsistencyimproveschainthought} and Self-Refine~\citep{self-refine}. Surprisingly, the prior prompting methods proposed before the reasoning LLMs era hurt the performance when the LLM itself is strong and emit long CoT out of the box, while ARQ is able to gain improvements with additional reasoning steps. 
However, ARQ slightly underperforms Self-Refine, which iteratively apply the LLM to solve and refine its previous solution. 
Our hypothesis is that the repeated-solving paradigm in the Self-Refine style methods benefits more directly from the existing RLVR~\citep{r1} pipeline than our relatively novel paradigm of asking useful questions. We also want to emphasize the methods on the right (MajVote and Self-Refine) are complementary to our approach as they can be applied to the solver within ARQ to get better performance. We demonstrate this compatibility in \cref{sec:app-with-majority-voting}. 
The result is encouraging, while also highlighting the limited improvement when applying ARQ on existing LLMs that may have not been trained to ask the right questions. Therefore, we describe a recipe for curating a dataset of good stepping stone questions and post-training in the next section.

\subsection{Data Curation and Post-training for ARQ}
\label{sec:method-train}

As we demonstrate in the previous section as well as in \cref{sec:good-stone}, existing LLMs generate a mixture of advantageous and detrimental stepping stones that lead to only moderate improvement on average. Therefore, a natural next step is to explore whether we can fine-tune LLMs to generate better stepping stones to achieve better performance under the ARQ framework. One option would be to collect annotated (\emph{Question}, \emph{CoT}, \emph{Stepping Stones}) tuples from expert humans. However, this approach is expensive and limited by the human ability to generate good stepping stone problems for LLMs. Instead, we study the effect of fine-tuning LLMs to ask the right questions via a synthetic data generation pipeline.

A key step in generating synthetic datasets is to accurately evaluate the quality of the generated data. We define the score of a stone, $z$, as the expected reward on the target problem, $x$, averaged over Monte Carlo rollouts of solutions to the generated stone, $y_{z}$ and the target problem, $y$; that is,
\begin{align}
    S(z,x) = \mathbb{E}_{y_{z} \sim \pi(z), y\sim \pi(z, y_z, x)}  R(x,y).
    \label{eq:stone-score}
\end{align}
Compared to the inference section where we roll out the solver $\pi$ twice to get solutions to the stepping stone and the target question, here we roll out the solver $2n$ times in total to estimate the score in~\cref{eq:stone-score} where $n$ is the sampling budget.

To generate synthetic data for fine-tuning, we start from a dataset of problems and use an existing LLM to generate a pool of candidate stepping-stone problems for each target problem. We evaluate each stone with $k$ rollouts to estimate its score $S(z,x)$, using an off-the-shelf LLM as the solver $\pi$. 
We then fine-tune the stone generator $\phi$ using a standard two-stage approach. First, we collect the highest-scoring stones for each target problem in the original dataset and use them for SFT. Next, we construct a preference dataset of paired stepping stones using the top- and bottom-performing ones, and use it to further fine-tune the LLM with DPO~\citep{rafailov2023direct}. Using these synthetically generated datasets, we substantially improve the performance of base LLMs as stone generators within the ARQ framework.
\section{Experiments}
\label{sec:experiments}

The experiment section is divided into the following parts. First recall that in~\cref{sec:limitation-inference-only}, we demonstrated that running ARQ on existing LLMs bring limited improvement compared to methods like Self-Refine when averaged over 20 individual runs where each run has its own stepping stone. In ~\cref{sec:good-stone}, we examine the score of individual stones generated with different LLMs as the stone generator. We show that existing LLMs generate a mixture of good and bad stepping stones, and that the good stepping stones are more helpful towards solving the target problem. Then, in \cref{sec:transfer}, we show that under our scoring system, the best stepping stones are not merely overfitting to a particular solver. Their benefit transfers to solvers of varying strength.
The strong benefit of good stepping stones and their transferability lay the foundation for the post-training experiments in \cref{sec:post-train}, where we fine-tune LLMs to generate better stepping stones. Finally, we extend ARQ to multiple stepping stones and evaluate two instantiations of this idea using both off-the-shelf LLMs and our fine-tuned LLMs.

\begin{figure*}[t]
  \centering
  \includegraphics[width=\textwidth]{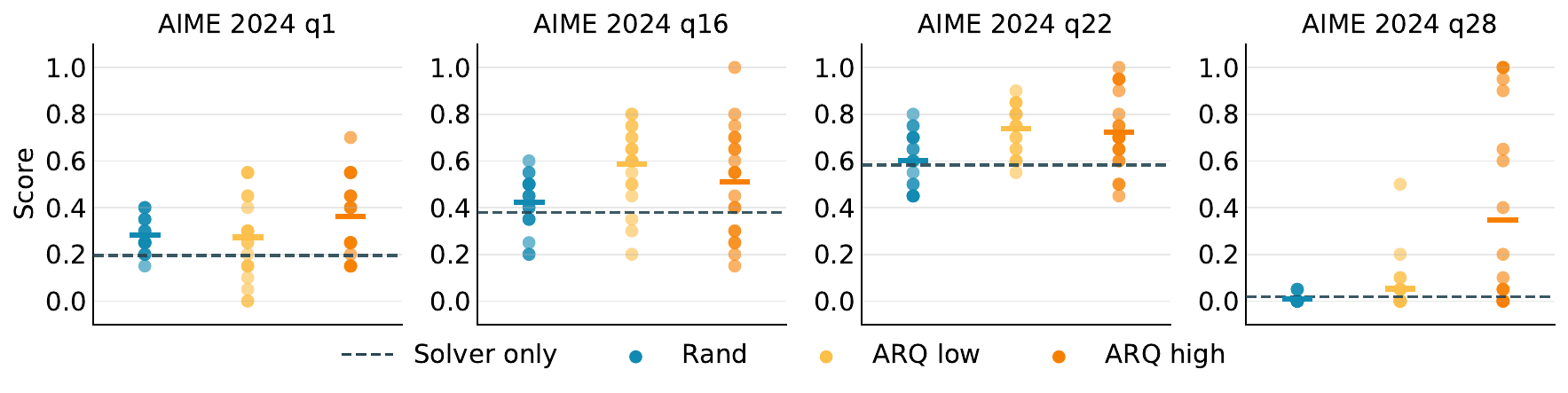}
  \vspace{-6mm}
  \caption{\small Scores of individual stones on selected questions from AIME 2024. Each point represents the score of the target problem conditioned on a generated stone, and the short horizontal bar indicates the average score over all stepping stones. The horizontal dashed line is the average performance of the \emph{Solver only} baseline. ARQ generates stepping stones of varying quality. The best ones are highly beneficial, but some are detrimental and reduce the average improvement. ARQ with the more capable reasoning LLM is able to generate better stones, both in terms of the best and the average.}
  \label{fig:all-stones}
  \vspace{-4mm}
\end{figure*}

\begin{figure}[!t]
  \centering
  \begin{minipage}[t]{0.49\textwidth}
    \vspace{0pt}
    \centering
    \includegraphics[width=\linewidth]{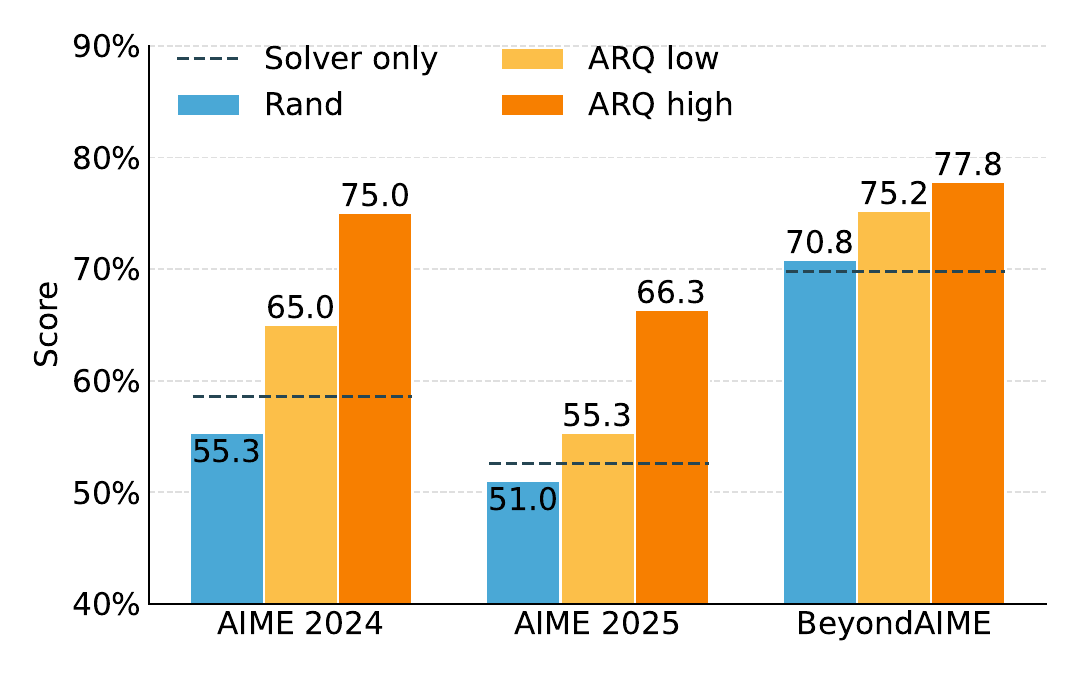}
    \captionof{figure}{\small The performance of ARQ variants conditioned on the \emph{best} stepping stone. Existing LLMs can generate highly useful stepping stones that increase the success rate on the target problems, and LLMs with better reasoning capabilities generate better stepping stones. In contrast, the best stone from \emph{Rand} does not improve over the \emph{Solver only} baseline (horizontal dashed lines). To avoid selection bias, we use half of the rollouts to select the best stones while we report the score using the other half.}
    \label{fig:stone-perf-best}
  \end{minipage}
  \hfill
  \begin{minipage}[t]{0.49\textwidth}
    \vspace{4pt}
    \centering
    \includegraphics[width=\linewidth]{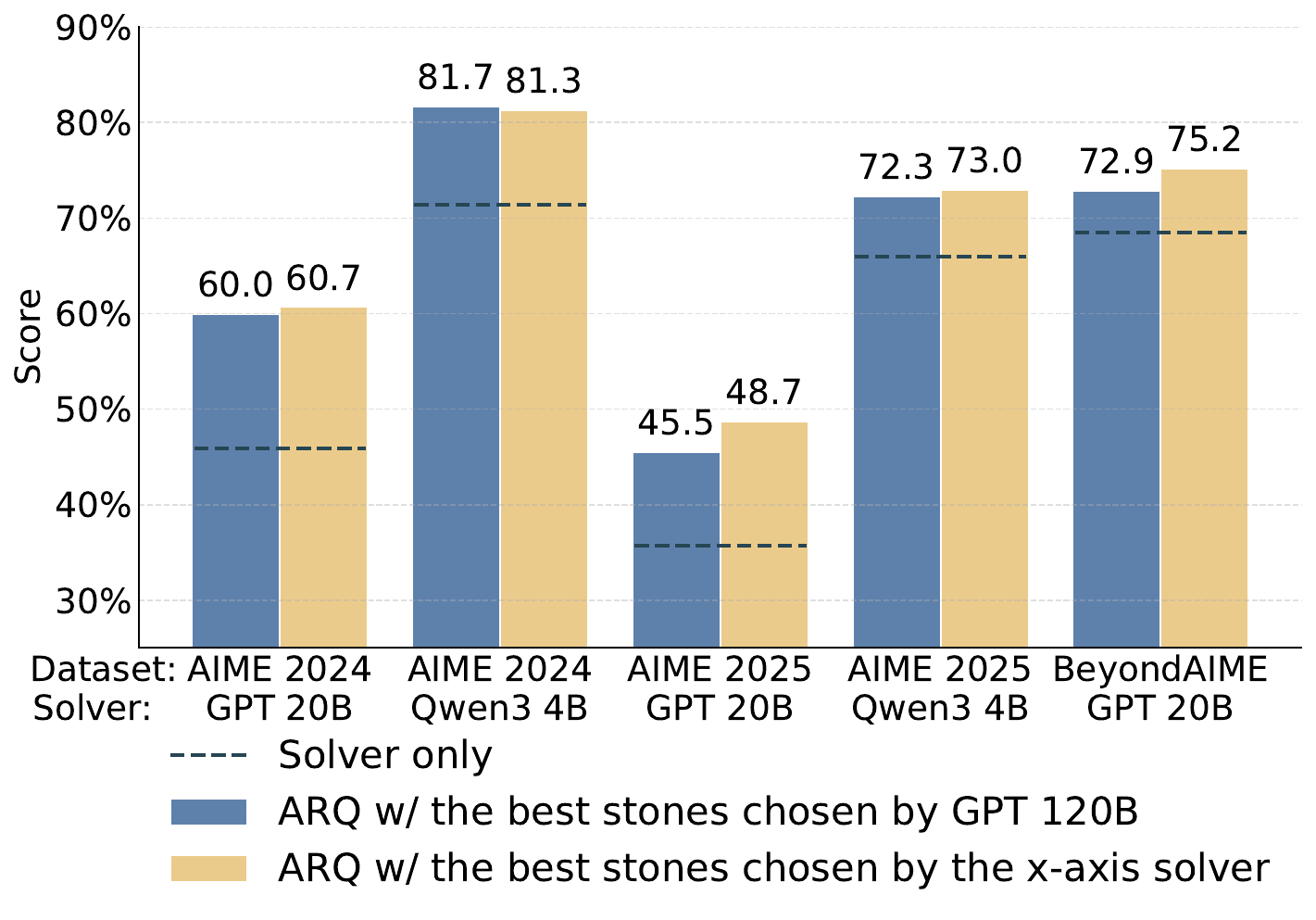}
    \captionof{figure}{\small Transferability results of good stepping stones. Performance remains roughly the same regardless of whether the best stones are selected by the new solvers themselves or by a reference solver (GPT 120B) that has a different size and reasoning capacity. This indicates that good stepping stones are universally helpful across solvers. The \emph{Solver only} scores of the new solvers are shown as dashed lines.}
    \label{fig:transfer-results}
  \end{minipage}
  \vspace{-4mm}
\end{figure}



\subsection{Delving into Individual Stepping Stones}
\label{sec:good-stone}


\textbf{Setup:} 
We use AIME 2024, AIME 2025, and BeyondAIME as our test benchmarks. The AIME 2024 and AIME 2025 datasets are widely used reasoning datasets. Each of them contains 30 challenging questions from the prestigious math competition for high school students. BeyondAIME~\citep{seed2025seed15thinkingadvancingsuperbreasoning-beyondaime} is a dataset of 100 questions that are much more challenging. We use MathVerify~\citep{Kydlicek_Math-Verify_Math_Verification} to compare generated answers against the ground truth.
For the solver LLMs, we use GPT-OSS-120B with low reasoning effort in AIME 2024 and AIME 2025 and the same model with high reasoning effort in BeyondAIME due to their different difficulty levels. 
Using these two levels of reasoning effort also helps us understand the performance of different methods when combined with LLMs of different capabilities.

For stepping stone generation, we consider a \textit{Rand} baseline and two \textit{ARQ variants}: 
\begin{itemize}[leftmargin=*, itemsep=0pt, parsep=0pt, topsep=2pt, partopsep=0pt]
    \item \textbf{\emph{Rand}}: a baseline that uses randomly generated stepping stones; the stone generator does not take the target problem as input and is asked to generate a random AIME-style question;
    \item \textbf{\emph{ARQ low}}: ARQ using GPT-120B with reasoning effort \textit{low} as stone generator $\phi$, simulating a naive generator with superficial thinking before asking questions;
    \item \textbf{\emph{ARQ high}}: ARQ using GPT-120B with reasoning effort \textit{high} as stone generator $\phi$, simulating the effect of having stepping stones from a more capable ``oracle''.
\end{itemize}
We first run the stone generator to generate 20 stepping stones independently. Then, to compute scores for each stone following~\cref{eq:stone-score}, we sample 20 rollouts per stone, i.e., 20 pairs of sampled solutions to the stepping stone and the target problem.

\textbf{Results:} We first look at the scores of individual stones \cref{fig:all-stones} on selected AIME 2024 problems. The full plots for all AIME 2024 and AIME 2025 problems are in \cref{fig:all-stone-aime24-appendix,fig:all-stone-aime25-appendix} of \cref{sec:appendix-fig-good-stone}. \cref{fig:all-stones} reveals that there is significant variance in the scores across generations. Many stones score much lower than the solver only baseline, even when using the more powerful LLM as the stone generator. Two possible factors contribute to the failure cases. One is that the generated stones are irrelevant or misleading, biasing the solver in the wrong direction. The other is that the generated questions are unsolvable for the solver and the wrong or invalid solutions in the context cause contamination when addressing the target problem. On the positive side, we notice that the best stepping stones are able to lead to significantly higher scores on the target problem, indicating that we may fine-tune LLMs to ask better questions with this score as a reward signal.

Next, we focus on the performance the best generated stepping stones averaged over the three datasets to get a holistic view. To report the scores of the best stepping stones, we use the average score of the first 10 solutions per stone to select the best stones and report their scores using the remaining 10 solutions to avoid selection bias. 
In \cref{fig:stone-perf-best}, we first notice that the best stepping stones substantially improve the performance in the ARQ framework. 
The best stones from \emph{ARQ low} improve the performance by $5\%$ over the \emph{solver only} baseline, while the ones from \emph{ARQ high} improve it by $13\%$. This improvement is not merely an effect of maximization over 20 runs or having more context, as it brings no improvement when maximizing over the 20 uninformative stepping stones in \emph{Rand}. 
Second, between the two ARQ versions, we see generating useful stepping stones could be an emergent behavior for off-the-shelf LLMs, with the more capable reasoning models generating much better stepping stones. For a more intuitive understanding of the best stepping stones and the behavior difference between \emph{ARQ high} and \emph{ARQ low}, please refer to \cref{sec:qualitative} for a qualitative analysis.

Obviously, the best score is not attainable at test time, as it requires access to the reward function in the math domain. However, this analysis is crucial to understand the strengths and weaknesses of ARQ on existing LLMs and paves the way for the post-training experiment in \cref{sec:post-train}.

\subsection{Transferability of Good Stepping Stones}
\label{sec:transfer}



After identifying the good stepping stones, the next question is whether the benefit from the good stones transfers and generalizes across different solvers. 

\textbf{Setup}:
The core idea for evaluating the transferability of good stepping stones is to take a set of \textit{best stones} selected by one solver (\emph{reference solver}) and evaluate them using a new solver (\emph{target solver}). Then we compare the resulting score against ARQ using the best stones selected by the \emph{target solver} itself. 
Specifically, we use the same datasets as in the previous section and use the stepping stones generated by \emph{ARQ high} as the candidate pool. The reference solvers are the same ones used in the previous experiments, i.e., GPT-120B low for AIME 2024 and AIME 2025, and GPT-120B high for BeyondAIME. 
For the new target solvers, we use GPT-20B, a smaller model, with low and high reasoning efforts for AIME and BeyondAIME respectively. We also consider a Qwen3-4B (thinking) solver on AIME to evaluate transferability between solvers of different families.
We use 20 rollouts per stone to score each stepping stone. When evaluating the stones selected by the target solvers themselves, we also perform the previously mentioned procedure to avoid bias, i.e., we use separate sets of rollouts for stone selection and evaluation.

\textbf{Results}: 
\cref{fig:transfer-results} shows the results. First, it is worth noting that the new solvers have noticeably different strengths from the reference solver in the previous experiment. For example, on the AIME 2024 and AIME 2025 datasets, the new \emph{GPT-20B low} solver is 21\% worse than the reference solver on average, while the \emph{GPT-20B medium} solver is 27\% better.
However, running ARQ with the best stepping stones improves the performance of all, reinforcing the observation that good stepping stones are helpful.
Regarding the transferability of good stepping stones, the performance of ARQ using the stones selected by the reference solvers is on par with that using the stones selected by the target solvers shown on the x-axis. This shows that good stepping stones are generally useful across solvers of different reasoning and problem-solving capacity. They improve performance not because they overfit to some specific biases of a particular solver but rather because they are genuinely useful questions to ask and solve first. 

\subsection{Fine-tuning LLMs to Generate Stepping Stones}
\label{sec:post-train}

Next, we show that we can improve LLMs' ability to ask better questions by fine-tuning them on the ARQ task following the process described in \cref{sec:method-train}.

\textbf{Setup:} We fine-tune Qwen3-8B (a thinking model) \citep{yang2025qwen3technicalreport} and Qwen2.5-32B-Instruct \citep{qwen2025qwen25technicalreport} to cover two representative starting points for additional post-training.
Qwen3-8B is a capable reasoning model that has been extensively post-trained on various reasoning tasks. This model performs decently on math benchmarks, but we find that it is not not good at generating useful stepping stones out of the box.
Qwen2.5-32B-Instruct is a base model that has not been post-trained for reasoning tasks. This model potentially has higher capacity, but it struggles to generate any stepping stones as it fails to follow our prompt and formatting instructions before fine-tuning.

\ifmeta
\begin{wrapfigure}[17]{r}{0.48\textwidth}
  \centering
  \vspace{-4mm}
  \includegraphics[width=\linewidth]{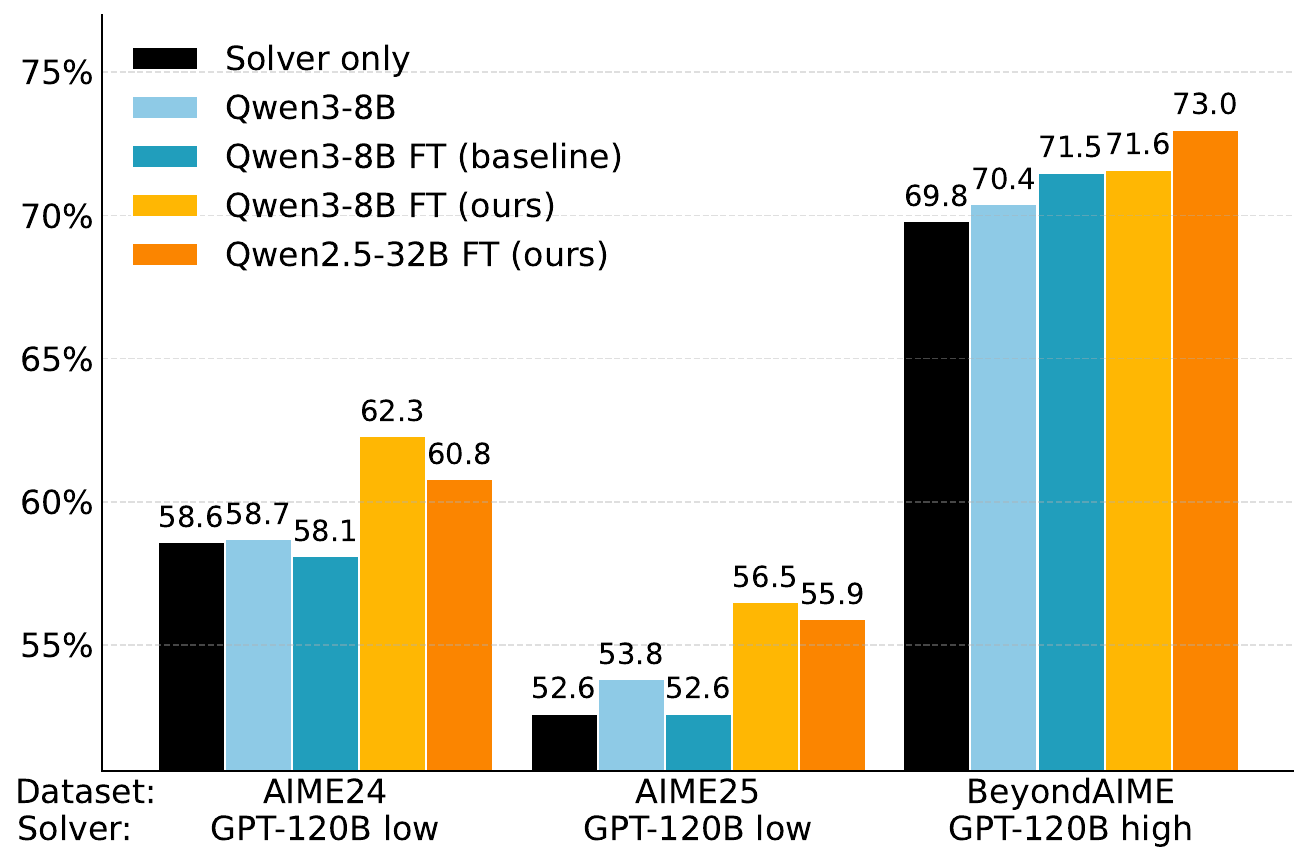}
  \caption{\small Performance of ARQ with Qwen3-8B and Qwen2.5-32B before and after fine-tuning (FT). Our fine-tuning procedure noticeably improves their performance as stone generators for different solvers.}
  \label{fig:ft-results}
\end{wrapfigure}
\else
\begin{wrapfigure}[16]{r}{0.48\textwidth}
  \centering
  \vspace{-4mm}
  \includegraphics[width=\linewidth]{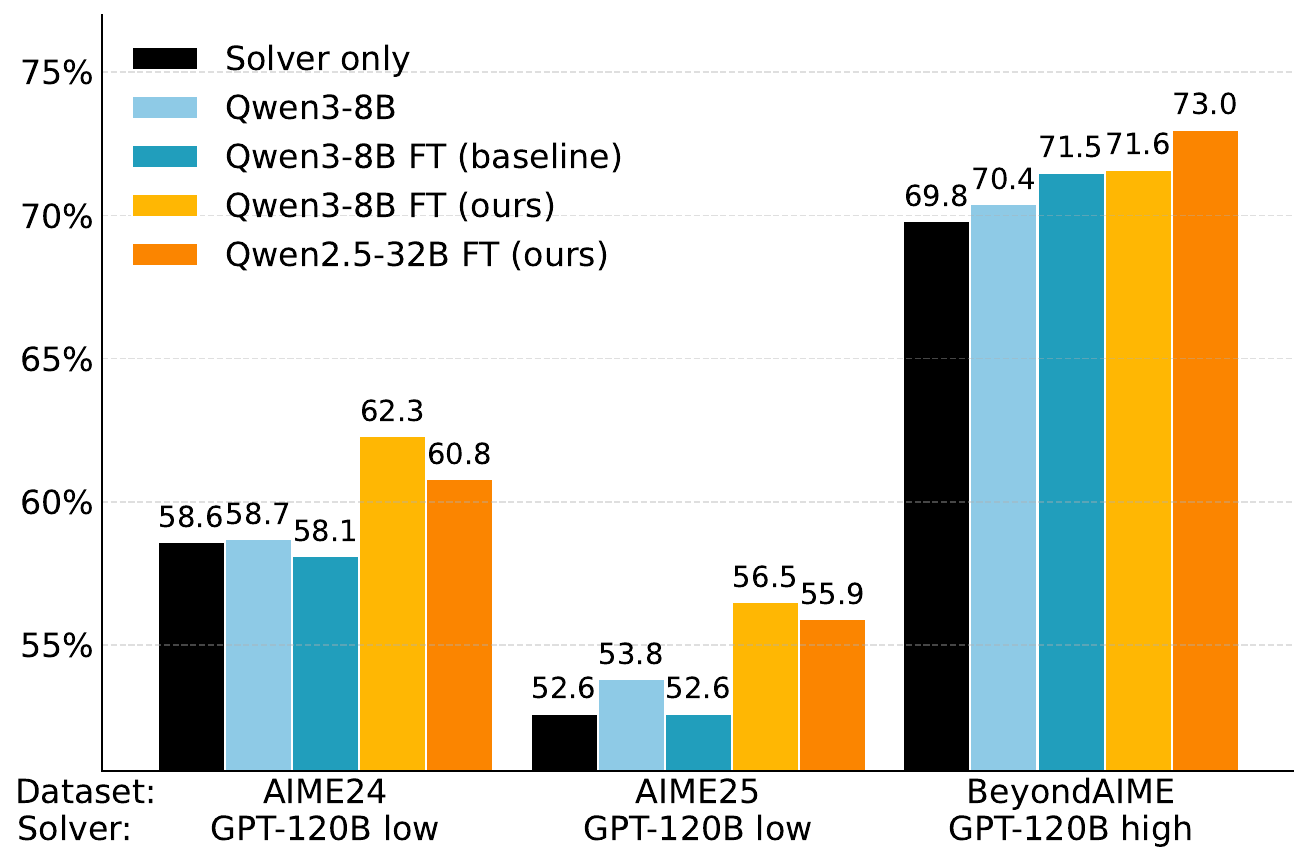}
  \caption{\small Performance of ARQ with Qwen3-8B and Qwen2.5-32B before and after fine-tuning (FT). Our fine-tuning procedure noticeably improves their performance as stone generators for different solvers.}
  \label{fig:ft-results}
\end{wrapfigure}
\fi

To collect the dataset for fine-tuning, we run ARQ with GPT-120B high as the stone generator and GPT-120B low as the solver on all 918 pre-2024 AIME questions. We generate 20 stepping stones per question and use our scoring function to assign a reward to each stepping stone. 
We first run SFT on the best stepping stones, followed by DPO on paired stones constructed from the pool of stones. Additional details of the data curation and SFT + DPO hyper-parameters are in \cref{sec:impl-detail}. To verify the importance of the scoring function and differentiate our procedure from pure distillation, we also consider a \textbf{baseline} where we repeat the exact fine-tuning procedure on the 8B model but the stones have randomly assigned scores.

We evaluate both LLMs before and after our fine-tuning procedure following the same practice as in \cref{sec:good-stone}.
We run the ARQ inference pipeline 20 times per problem and report the \emph{average} performance as an indicator of whether the models have improved their question-asking capability in general. As before, we use GPT-120B low as the solver in AIME 2024 and AIME 2025, and GPT-120B high as the solver in BeyondAIME. 

\textbf{Results:}
\cref{fig:ft-results} summarizes the results. We omit the score of Qwen2.5-32B as the stone generator before fine-tuning as it fails to follow our prompt. The Qwen3-8B reasoning model is able to generate well-formatted stepping stones, but they only improve performance by $0.6\%$ on average compared to the solver-only baseline. After fine-tuning, both models greatly improve their question-asking capability, leading to improvements of $3.1\%$ for Qwen3-8B and $2.9\%$ for Qwen2.5-32B averaged over three datasets. The models now generate questions that are helpful both for a weak solver (as in AIME 2024 and AIME 2025) and for a solver that is significantly stronger than themselves in terms of model size and reasoning capability (as in BeyondAIME). 
The 8B model fine-tuned with the baseline approach, however, has mixed results across the three datasets. It performs noticeably worse than 8B (ours) on the two AIME datasets but on par on the hard BeyondAIME dataset, potentially because the strong solver used in BeyondAIME becomes the more dominant factor for performance. This result is particularly encouraging given the relatively small amount of data used for fine-tuning. It further validates the potential of stepping stone generation as a promising and interesting post-training task.

\subsection{Extending ARQ to Multiple Stones}
\label{sec:more-stones}

\begin{figure*}[t]
  \centering
  \begin{tabular}{@{}c@{\hspace{0.00\textwidth}}c@{}@{\hspace{0.00\textwidth}}c@{}}
    \includegraphics[width=0.33\textwidth]{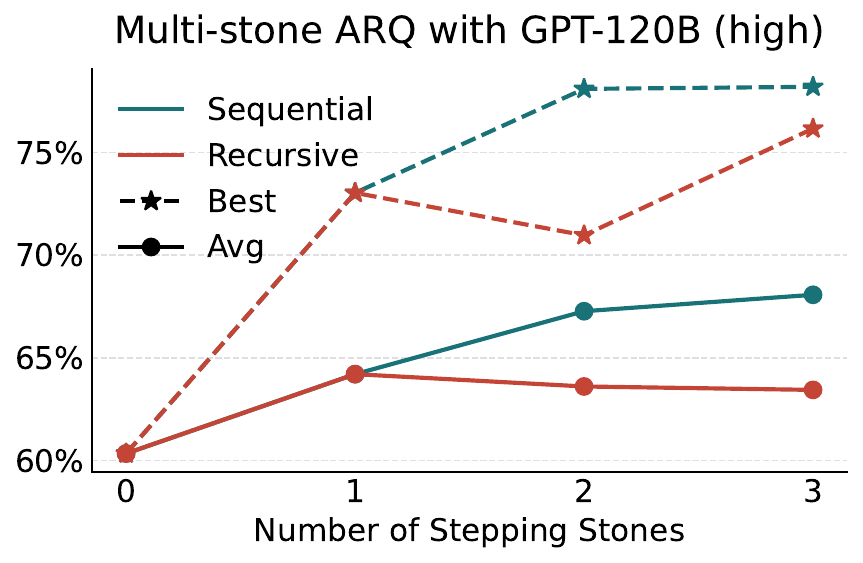} &
    \includegraphics[width=0.33\textwidth]{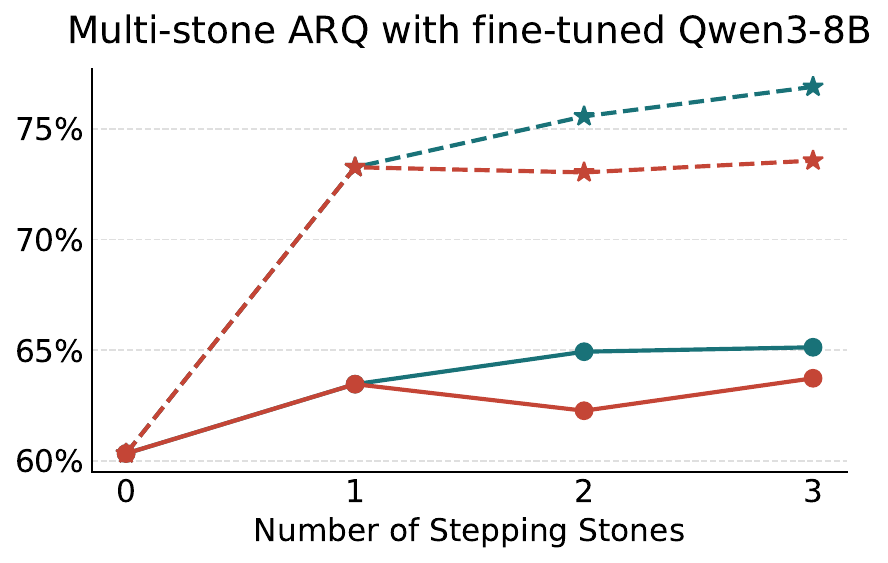} & 
    \includegraphics[width=0.33\textwidth]{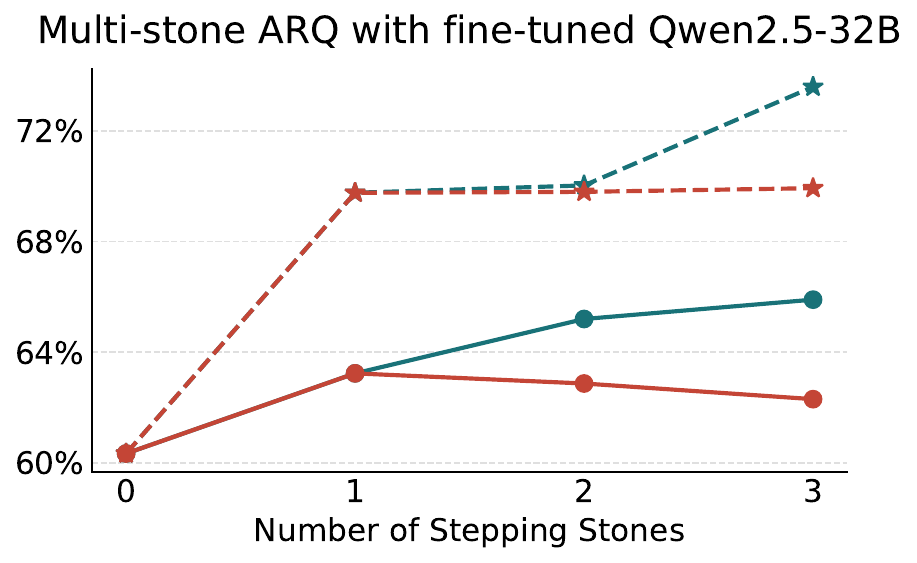}
  \end{tabular}
  \caption{\small Performance of ARQ with multiple stones. We run ARQ with GPT-120B high (left), fine-tuned Qwen3-8B (middle), and fine-tuned Qwen2.5-32B (right) as stone generators. Results are averaged over the AIME 2024, AIME 2025, and BeyondAIME benchmarks. Each stone generator is asked to produce 20 sets of either 2 or 3 stepping stones, generated either \emph{sequentially} or \emph{recursively}. \textbf{Best} denotes performance conditioned on the best-performing set of stones, while \textbf{Avg} denotes performance averaged over the 20 sets. \emph{Sequentially} generated stepping stones continuously improve both best-case and average-case performance. Results with 0 stepping stones correspond to the \textit{solver only} in the previous sections.}
  \label{fig:multi-stone}
  \vspace{-6mm}
\end{figure*}

So far, we have focused on ARQ with a single stepping stone. Next, we evaluate whether increasing the number of stepping stones brings further improvement. 

\textbf{Setup:}
We consider two strategies for extending ARQ to multiple stepping stones discussed in \cref{sec:method-inference}: generating stepping stones \emph{sequentially} or \emph{recursively}.
In the sequential case, the goal is to gradually build a curriculum, where each new question aims to bridge the gap between the existing stepping stones and the target problem. In the recursive case, each new stepping stone is generated to help solve the previously generated stepping stone, treating it as the new target problem. The prompts for sequential and recursive stone generation are listed in \cref{sec:appendix-prompt}.

We evaluate both strategies on the same three math benchmarks used in previous experiments, using the same solvers as in \cref{sec:good-stone} for each benchmark. For the stepping stone generators, we consider an off-the-shelf model (GPT-120B with high reasoning effort) as well as the two fine-tuned LLMs from \cref{sec:post-train}, Qwen3-8B FT (ours) and Qwen2.5-32B FT (ours). Note that sequential stone generation is a generalization test for our fine-tuned models, because they have not seen prompts in which they are given both the target problem and one or more existing stepping stones during our fine-tuning stage.

We run both multi-stone generation strategies to produce 20 sets of either 2 or 3 stepping stones for each benchmark problem. For each set, we use the solver to solve each stone and then solve the target problem, conditioning on the available stones and their solutions as context. We repeat this procedure 20 times to estimate the average score for each set of stones.

\textbf{Results:}  
\cref{fig:multi-stone} summarizes the results for each model averaged over all three benchmarks. We also provide the separate results for each model on each dataset in \cref{sec:appendix-multiple}.
The results show that sequential ARQ leads to substantial gains in both average and best performance across all three LLM backbones, while the recursive approach overall exhibits fluctuation or degradation.
For the sequential version in particular, we observe a $5.2\%$ absolute improvement in the best score and $3.9\%$ in the average score when scaling from one stepping stone to three stepping stones using GPT-120B high as the stone generator. 
Notably, this performance gain extends to models that have only been fine-tuned on single stepping stone generation, demonstrating additional generalization benefits from the ARQ post-training procedure. The fine-tuned Qwen3-8B and the fine-tuned Qwen2.5-32B achieve absolute improvements of $1.7\%$ and $2.7\%$ in the average score, respectively, and their best scores improve by  $3.7\%$ and $3.8\%$ respectively.

\section{Conclusion}
\label{sec:conclusion}
In this paper, we study LLMs’ ability to address hard problems by asking the right stepping-stone questions.
We propose the ARQ framework, which includes both an inference-time procedure to generate and utilize stepping-stone problems as well as a data-curation recipe and post-training procedure to fine-tune LLMs to generate better stepping stones.
We show that good stepping stones improve the chance of solving the target problem, and we demonstrate that their helpfulness generalizes to solvers of varying capabilities. Finally, we curate a stepping-stone dataset and show that we can post-train multiple LLMs to become better stepping stone generators.
The primary limitation of this work is the relatively small scale of the post-training experiment given constraints in computation and limited sources of hard target problems. The concept of stepping stones could be more useful in extremely challenging domains such as IMO level math and scientific research. Directions for future research include scaling up the study to larger datasets and other domains such as coding, as well as exploration of post-training LLMs to ask better questions via online RL.

\bibliographystyle{assets/plainnat}
\bibliography{reference}

\clearpage
\newpage
\beginappendix

\section{Qualitative Analysis for Stepping Stone}
\label{sec:qualitative}

\begin{table}[h]
\centering
\label{tab:beyond-aime-results}
\begin{tabular}{lccccc}
\toprule
Model & \makecell[c]{Scale\\ Reduction} & \makecell[c]{Constraint\\ Simplification} & Generalization & \makecell[c]{Intermediate\\ Step} & \makecell[c]{Method \\Practice} \\
\midrule
\emph{ARQ high} & 68 & 13 & 8 & 8 & 3 \\
\emph{ARQ low}  & 62 & 22 & 8 & 5 & 3 \\
\bottomrule
\end{tabular}
\vspace{1mm}
\caption{\small Distribution of the best stepping stones for 100 target problems in BeyondAIME among five categories. }
\label{tab:qualitative}
\end{table}

In \cref{sec:experiments}, we perform a series of experiments to understand the properties of stepping stones and their effect on solving target problems. 
To have a more intuitive understanding of how the stepping stone assists the solver, we manually review the stepping stone problems that produce the best roll-out performance for all 100 problems in BeyondAIME benchmark. We find that the stepping stones can be classified into five categories based on their strategies.
\begin{itemize}
    \item \textbf{Scale Reduction:} The stepping stone keeps the original question structure unchanged and only reduces the large parameters to small values that can be calculated manually;
    \item \textbf{Constraint Simplification:} The stepping stone freezes a free variable, reduces the dimension of the problem, or only considers a special case.
    \item \textbf{Generalization:} The stepping stone changes a large value to a generic parameter such as ``m'' or ``n'' and asks for a generic formula.
    \item \textbf{Intermediate Step:} The stepping stone figures out the prerequisite subtasks or critical lemmas that must be completed first.
    \item \textbf{Method Practice:} The stepping stone changes the question structure to practice the core math tricks involved in the question.
\end{itemize}

The distribution of the best-performing stepping stones for \emph{ARQ high} and \emph{ARQ low} among the five distributions is shown in \cref{tab:qualitative}. We could observe from the table that \emph{ARQ high} and \emph{ARQ low} exhibit a highly similar trend. Under both settings, most of the best-performing stepping stones fall into the category of scale reduction and constraint simplification, while only very few best-performing stepping stones work by teaching the solver the required math tricks.

\section{Compatibility with Inference Approach that Improves the Solver}
\label{sec:app-with-majority-voting}
\begin{table}[]
\centering
\setlength{\tabcolsep}{12pt} 
\setlength{\aboverulesep}{0.6ex} 
\setlength{\belowrulesep}{0.6ex} 
\begin{tabular}{lccc}
\toprule
            & \multicolumn{1}{l}{No Maj} & \multicolumn{1}{l}{Maj@20} & \multicolumn{1}{l}{Gain} \\
\midrule
\emph{Solver only} & 69.8                       & 76.0                         & +6.2                      \\
\emph{ARQ high}    & 72.8                       & 79.0                       & +6.2                      \\
\emph{ARQ low}     & 72.7                       & 77.0                       & +4.3   \\
\bottomrule
\end{tabular}
\vspace{1mm}
\caption{The performance of ARQ and \emph{solver only} baseline when combined with major voting.}
\label{tab:compatibility}
\end{table}

In \cref{sec:limitation-inference-only}, we compare our proposed ARQ with other similar methods that generate intermediate steps and two representative test-time scaling methods that directly operate on the solver, namely majority voting and self-refine. Notably, ARQ improves the problem-solving ability by introducing stepping stones with question generation $\phi$. 
Being agnostic to the problem solver $\pi$, ARQ is compatible with test-time scaling approaches that enhance the performance of solver $\pi$.  Here we use majority voting as an example, and the experimental results are in \cref{tab:compatibility}. As shown in the table, both \emph{ARQ high} and \emph{ARQ low} exhibit substantial gains with Maj@20, suggesting that ARQ can be combined with other test-time scaling approaches to achieve better performance.

\section{Supplementary Figures for \cref{sec:good-stone}}
\label{sec:appendix-fig-good-stone}

\begin{figure}
    \centering
    \includegraphics[width=\linewidth]{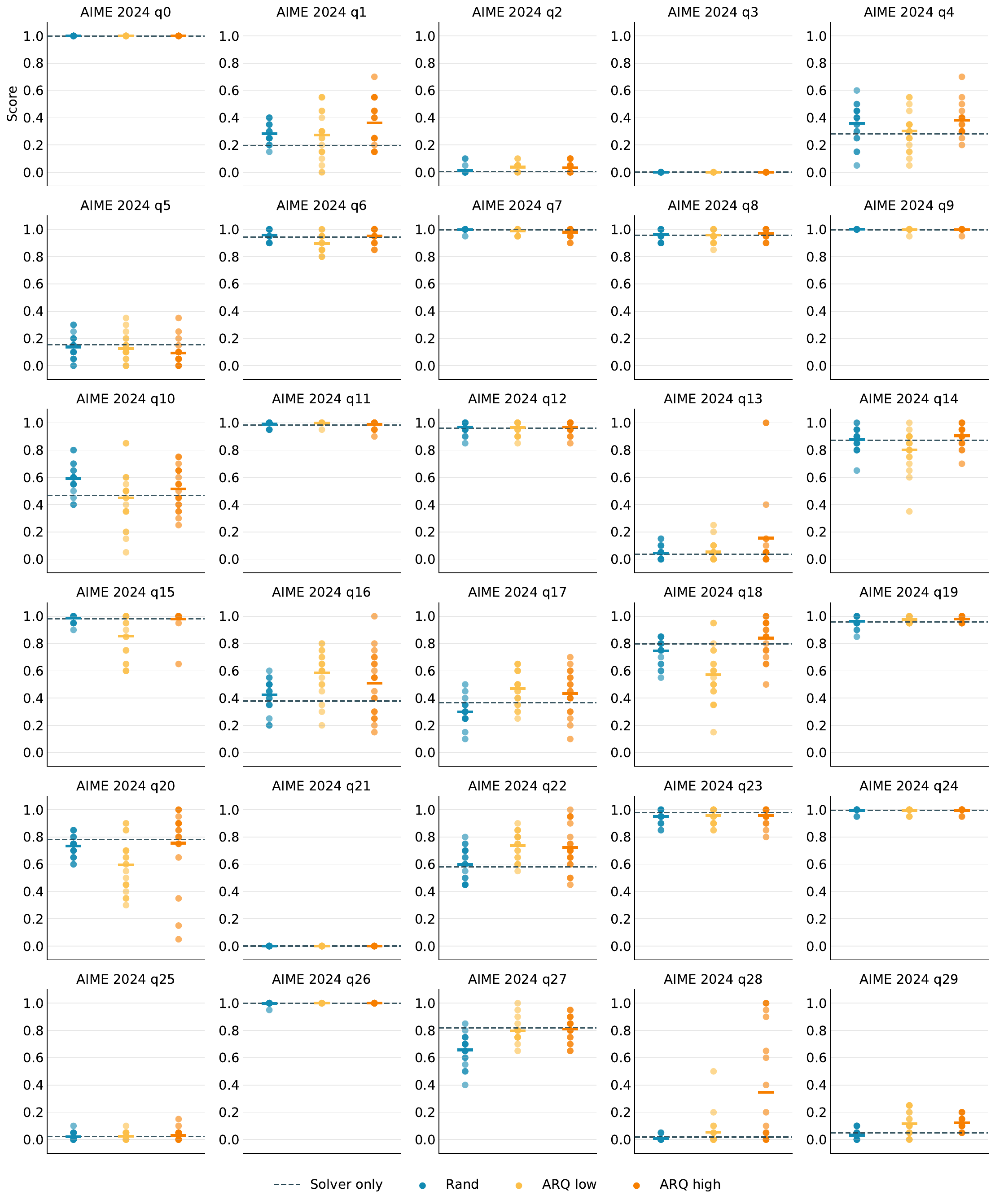}
    \caption{Score of individual stones on all questions from AIME 2024, a complete version of \cref{fig:all-stones}. The solver is GPT-120B with low reasoning effort for all methods.}
    \label{fig:all-stone-aime24-appendix}
\end{figure}

\begin{figure}
    \centering
    \includegraphics[width=\linewidth]{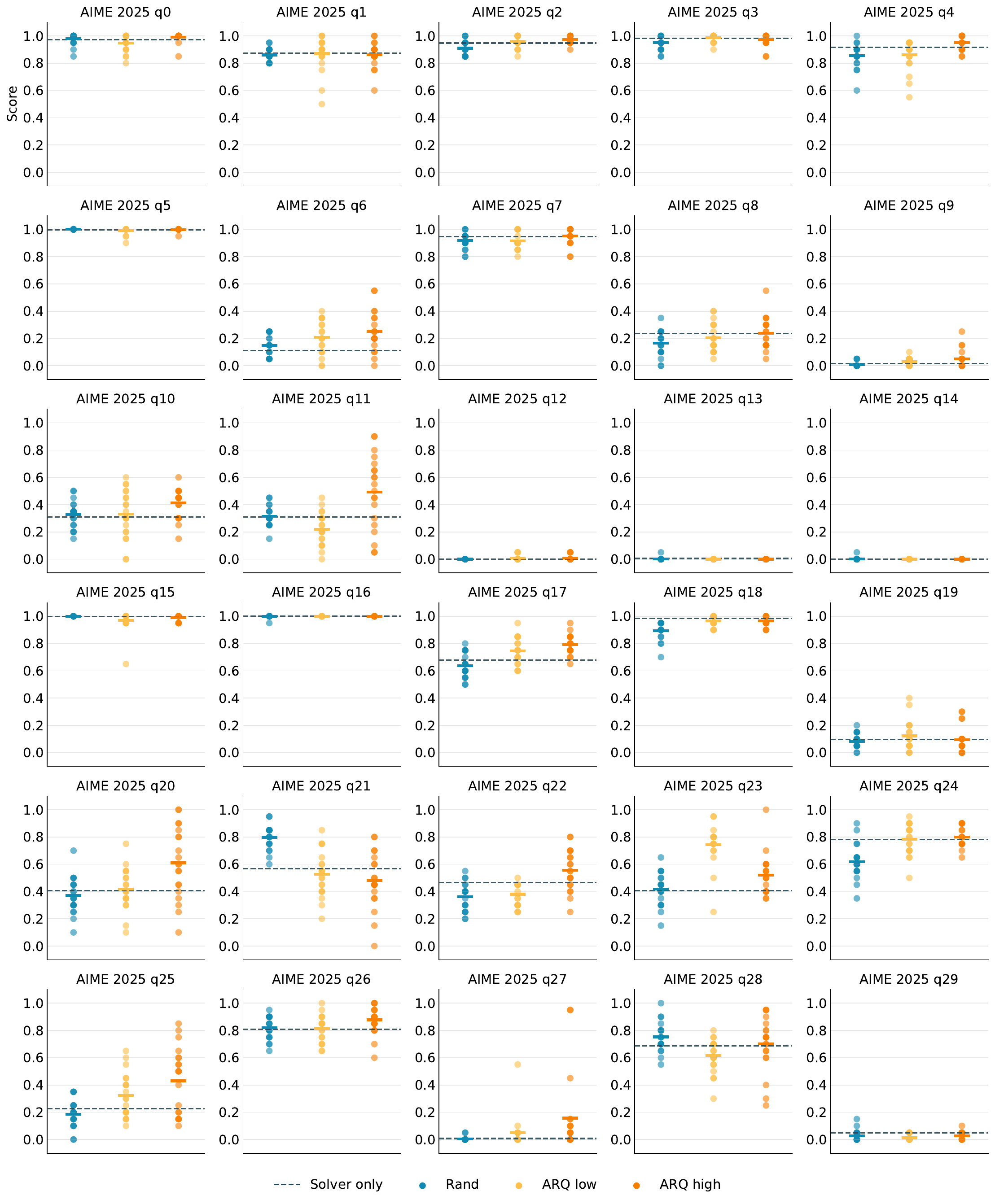}
    \caption{Score of individual stones on all questions from AIME 2025. The solver is GPT-120B with low reasoning effort for all methods.}
    \label{fig:all-stone-aime25-appendix}
\end{figure}

In \cref{fig:all-stones} of \cref{sec:good-stone}, we plot the performance of ARQ conditioned on individual stepping stones, together with a \emph{Rand} baseline that generates random stepping stones without conditioning on the target problems. In \cref{fig:all-stone-aime24-appendix} and \cref{fig:all-stone-aime25-appendix}, we show the results for all problems in AIME 2024 and AIME 2025. This holistic view reaffirms the observation that running ARQ with existing LLMs generates a diverse set of stepping stones containing both beneficial ones and detrimental ones. The LLM with better reasoning capability tends to generate more beneficial stepping stones with larger improvements. Both benchmarks also contain a number of questions where the solver itself can reliably solve the problem, in which case running ARQ is technically not necessary.

\section{Details of Dataset and Implementation of Post-Training in \cref{sec:post-train}}
\label{sec:impl-detail}

\begin{table}[h]
\centering
\small
\label{tab:dataset-statistics}
\begin{tabular}{lcccc}
\toprule
& Dataset Size & Mean Prompt Length & Mean CoT Length & Mean Stepping Stone Length \\
\midrule
SFT & 918          & 184.25          & 4034.45    & 66.24        \\
DPO & 1063         & 193.93          & 4911.18    & 74.54       \\
\bottomrule
\end{tabular}
\vspace{1pt}
\caption{\small The statistics of the SFT dataset and the DPO dataset in our experiments.}
\end{table}

\begin{figure}[h]
    \centering
    \begin{minipage}[t]{0.47\textwidth}
        \centering
        \includegraphics[width=\linewidth]{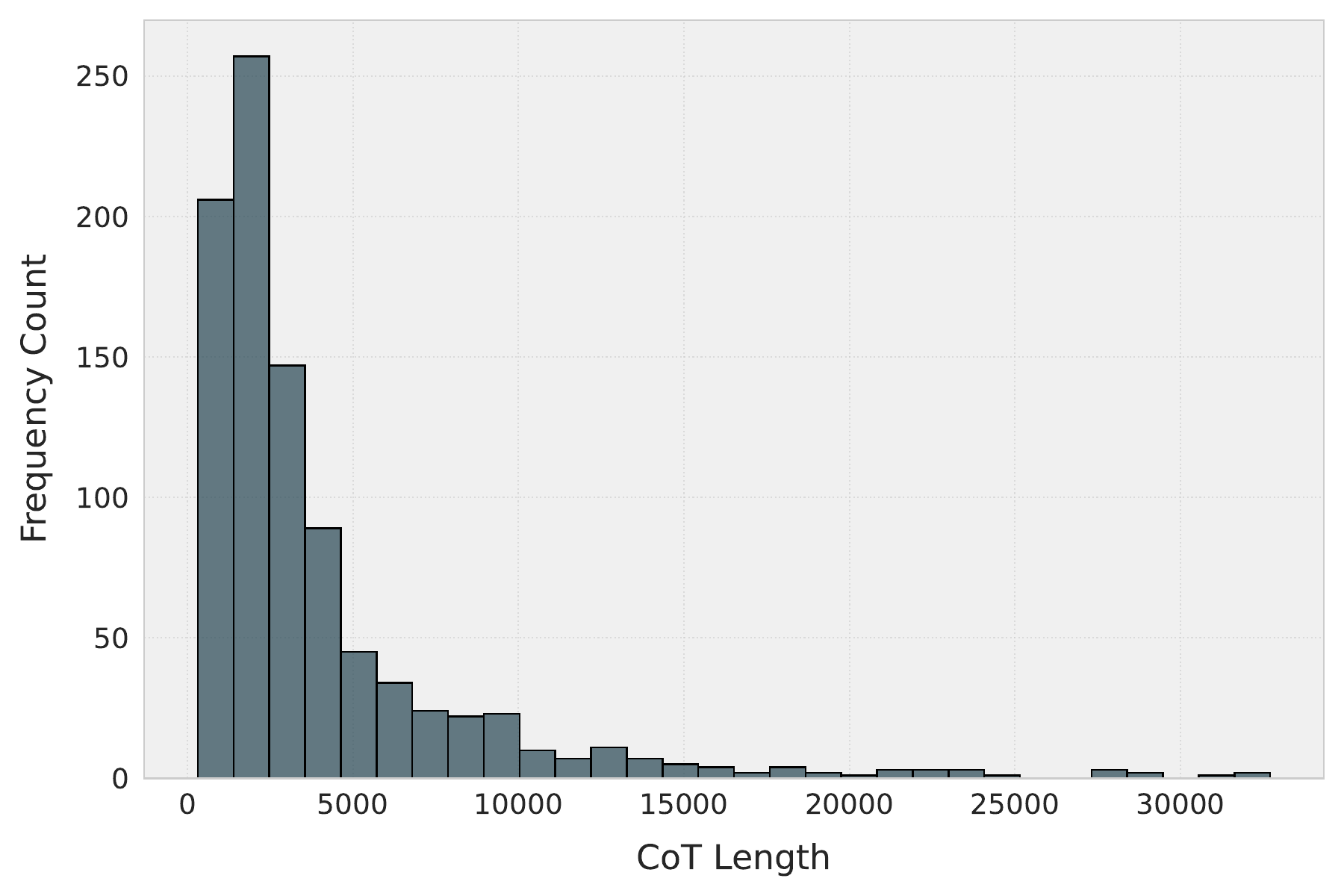}
    \end{minipage}\hfill
    \begin{minipage}[t]{0.47\textwidth}
        \centering
        \includegraphics[width=\linewidth]{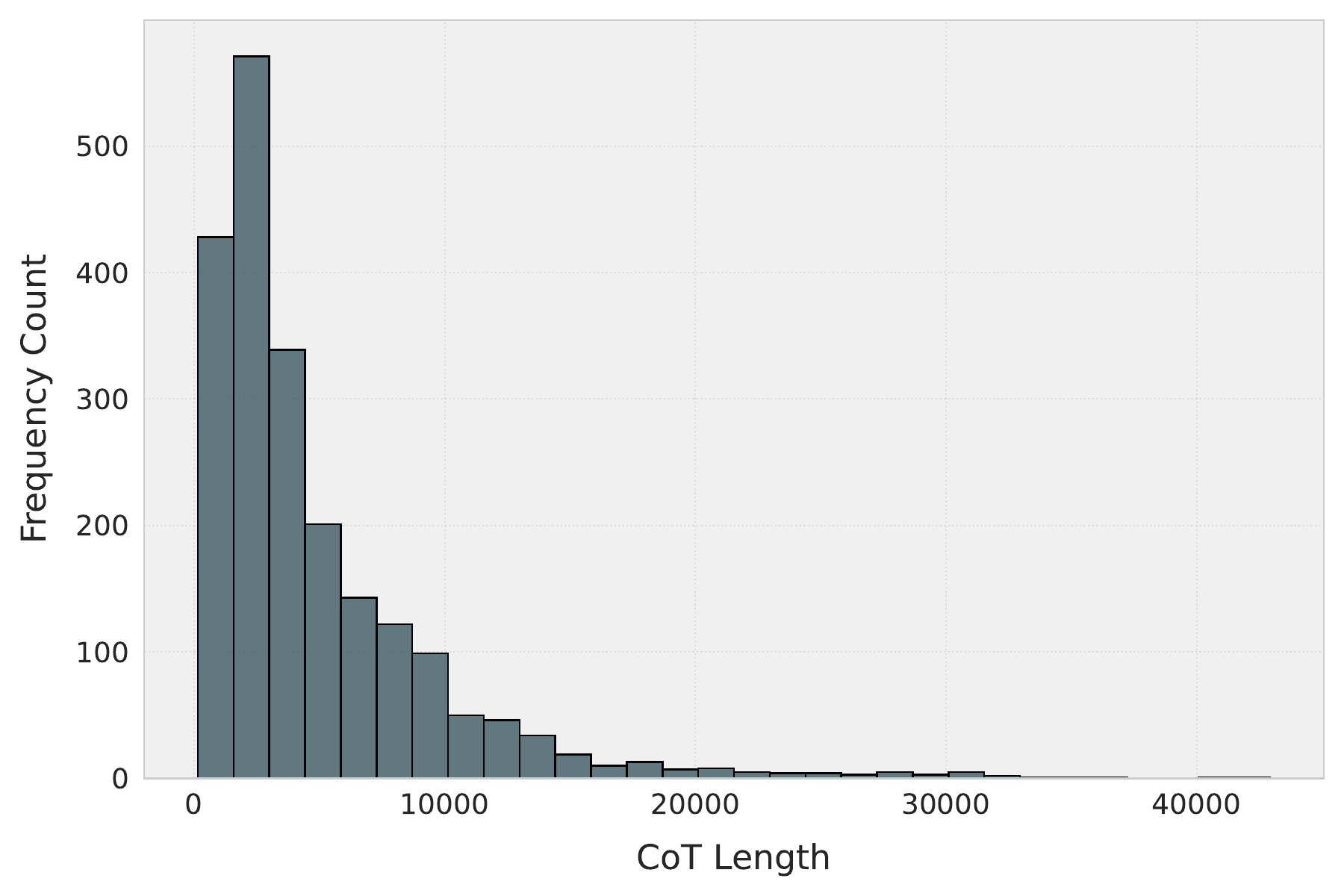}
    \end{minipage}
    \caption{The distribution of the CoT length in our constructed SFT dataset (left) and DPO dataset (right).}
    \label{fig:cot-length-distribution}
\end{figure}

In this section we elaborate on the data collection process and the implementation details for our our post-training experiments in \cref{sec:post-train}. 

\subsection{Data Collection}
\label{sec:appendix-dataset}

We take the 918 pre-2024 AIME questions as our source of target problems for both datasets. We first generate $20$ stepping stones for every target AIME problem using GPT-120B high as the stone generator. Then for each stepping stone we sample $20$ rollouts following \cref{eq:stone-score} to evaluate its quality. The stepping stone with the best score is included in the SFT dataset. 

Next, we take the $20$ generated stepping stones per problem to construct the DPO dataset. We rank the stepping stones by their scores and only keep the top-3 stones $z_1,z_2,z_3$ as well as the bottom-3 stones $z_{18},z_{19},z_{20}$. With the $6$ remaining stones, we construct three pairs of stones, namely $(z_{1},z_{20}), (z_2,z_{19})$ and $(z_3,z_{18})$. We compute the score difference for all pairs of stones and keep the ones with a difference $\geq 0.25$ in the DPO dataset. In principle, the rationale behind the data construction procedure is to 1) avoid having too many pairs from the same problem; 2) avoid having the same stone appear multiple times, which may cause overfitting; and 3) avoid using pairs where the score gap between the preferred and non-preferred is small. This procedure results in a dataset of 1063 preference pairs generated from 436 unique target problems. The average score difference over the selected pairs is $0.47$.

The statistics of the SFT dataset and the DPO dataset, including the dataset size and the mean length of prompts, chains of thought, and generated stones, are displayed in \cref{tab:dataset-statistics}. Note that for the DPO dataset, the length is averaged over both positive examples and negative examples.
Moreover, we plot the distribution of the CoT length in \cref{fig:cot-length-distribution}. 
From the figure, we observe that there is a long tail in the CoT length distribution, especially for the DPO dataset, where the CoT can be longer than 40,000.

\subsection{Implementation and Hyper-Parameters}

\label{sec:appendix-post-train-impl}

\begin{table}[t]
\centering

\begin{minipage}[t]{0.45\textwidth}
  \centering
  \begin{tabular}{lc}
    \toprule
    \textbf{Hyper-parameter} & \textbf{Value} \\
    \midrule
    \texttt{max\_length}      & 16384 \\
    \texttt{batch\_size}      & 16 \\
    \texttt{lr}               & 1e-5 \\
    \texttt{adam\_betas}      & (0.9, 0.95) \\
    \texttt{grad\_clip}       & 1.0 \\
    \texttt{weight\_decay}    & 0.0001 \\
    \texttt{warmup\_ratio}    & 0.05 \\
    \texttt{schedule}         & cosine \\
    \texttt{epoch}            & 5 \\
    \bottomrule
  \end{tabular}
  \label{tab:hyper_parameter_sft}
\end{minipage}\hfill
\begin{minipage}[t]{0.45\textwidth}
  \centering
  \begin{tabular}{lc}
    \toprule
    \textbf{Hyper-parameter} & \textbf{Value} \\
    \midrule
    \texttt{max\_length} & 14336 \\
    \texttt{batch\_size}           & 16 \\
    \texttt{lr}                    & 1e-6 \\
    \texttt{adam\_betas}           & (0.9, 0.95) \\
    \texttt{grad\_clip}            & 1.0 \\
    \texttt{weight\_decay}         & 0.01 \\
    \texttt{schedule}              & constant \\
    \texttt{epoch}                 & 1 \\
    \texttt{dpo\_beta}             & 0.1 \\
    \bottomrule
  \end{tabular}
  \label{tab:hyper_parameter_rl}
\end{minipage}
\vspace{10pt}
\caption{The value of the hyper-parameters in our reasoning-oriented training experiment (Section 5.2) for SFT (left) and RL (right).}
\label{tab:hyper_parameter}
\end{table}

We use Qwen3-8B~\citep{yang2025qwen3technicalreport} and Qwen2.5-32B-Instruct~\citep{qwen2025qwen25technicalreport} as the base models for post-training, but in principle the post-training approach is agnostic to specific language models. 
Our experiments are conducted on a cloud Linux server with Ubuntu 22.04.5 operating system. The codes are written in Python 3.12.8. We run our experiments on Nvidia H200 GPU.
The hyperparameters for SFT and DPO are shown in \cref{tab:hyper_parameter_sft}. 
Notably, we set the maximum sequence length in both experiments to be $16,384$, which is much longer than the average CoT length to include a small amount of extremely long sequences shown in \cref{fig:cot-length-distribution}. Such a large sequence length makes it difficult to perform full-parameter fine-tuning on DPO, since DPO requires an additional reference model to compute KL-divergence, and it needs to process both positive and negative examples. Therefore, we use LoRA~\citep{hu2022lora} for parameter-efficient fine-tuning with hyper-parameter $r = 256$ and $\alpha = 32$, following the setting of \citet{schulman2025lora}.

\section{Supplementary Figures for \cref{sec:more-stones}}
\label{sec:appendix-multiple}

\begin{figure}[t]
\centering

\begin{subfigure}{\textwidth}
  \centering
  \begin{minipage}[t]{0.32\textwidth}
    \centering
    \includegraphics[width=\linewidth]{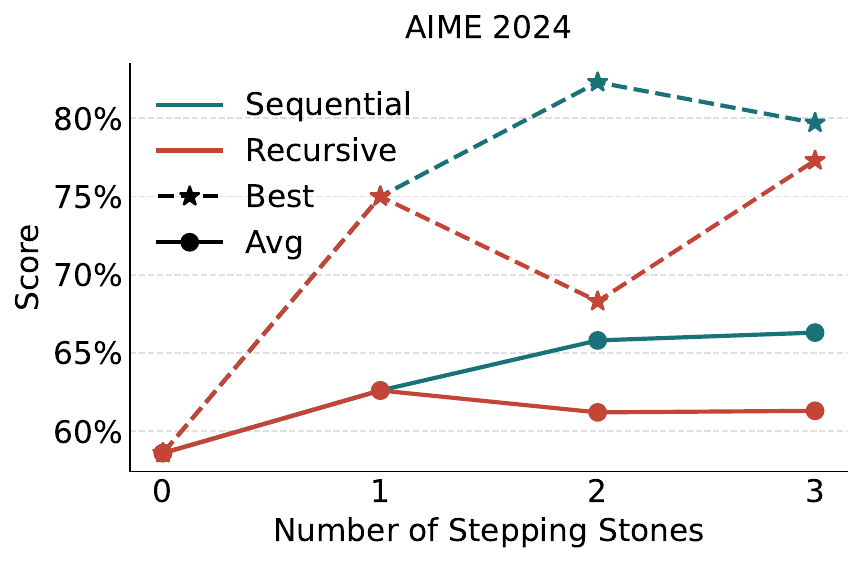}
  \end{minipage}\hfill
  \begin{minipage}[t]{0.32\textwidth}
    \centering
    \includegraphics[width=\linewidth]{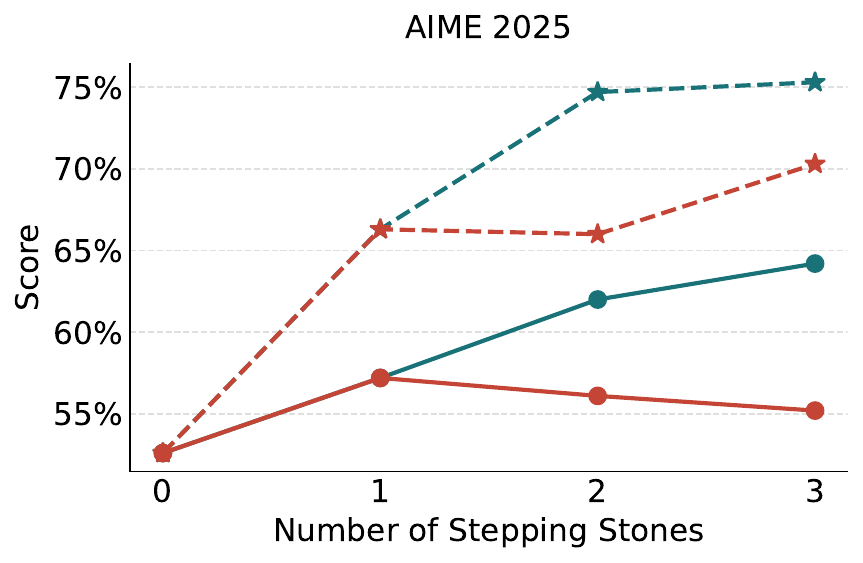}
  \end{minipage}\hfill
  \begin{minipage}[t]{0.32\textwidth}
    \centering
    \includegraphics[width=\linewidth]{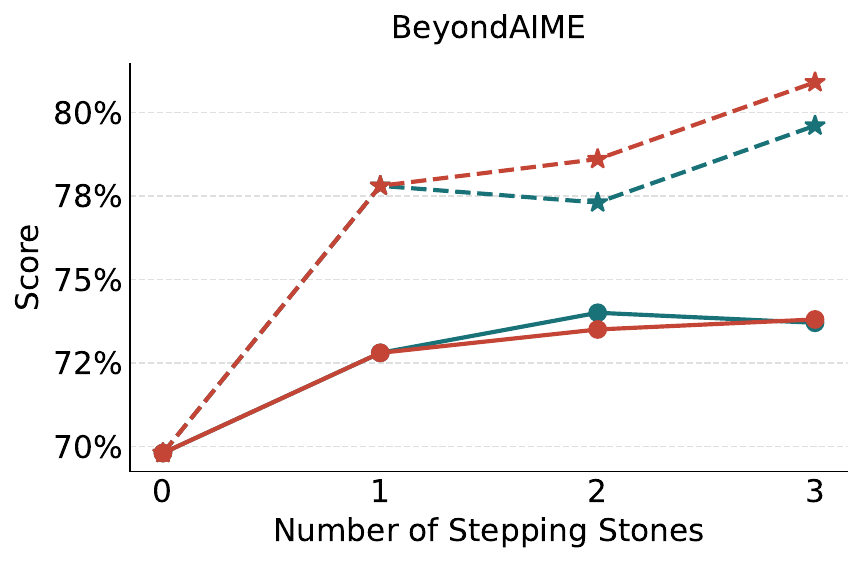}
  \end{minipage}

  \caption{\small Multi-stone ARQ with GPT-120B (high) as stone generator.}
  \label{fig:multistone-dpo-gpt120b}
\end{subfigure}

\vspace{0.8em}

\begin{subfigure}{\textwidth}
  \centering
  \begin{minipage}[t]{0.32\textwidth}
    \centering
    \includegraphics[width=\linewidth]{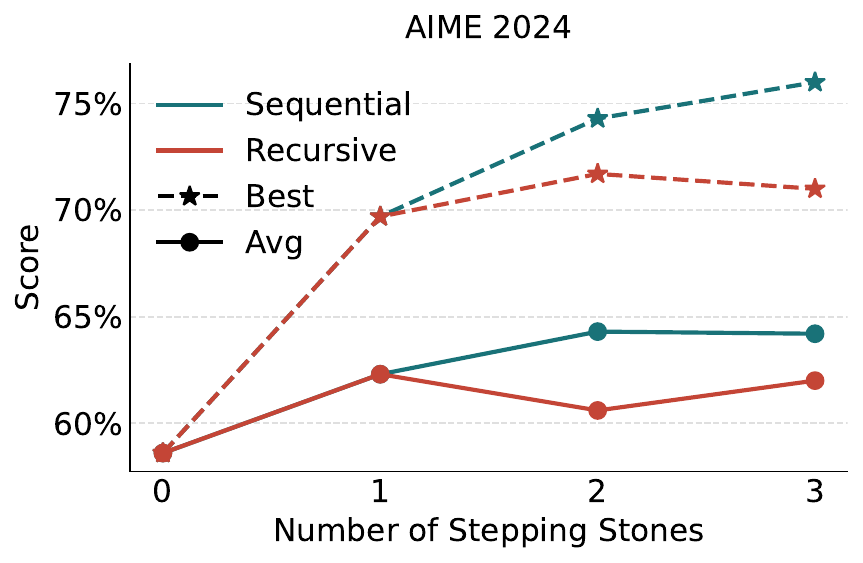}
  \end{minipage}\hfill
  \begin{minipage}[t]{0.32\textwidth}
    \centering
    \includegraphics[width=\linewidth]{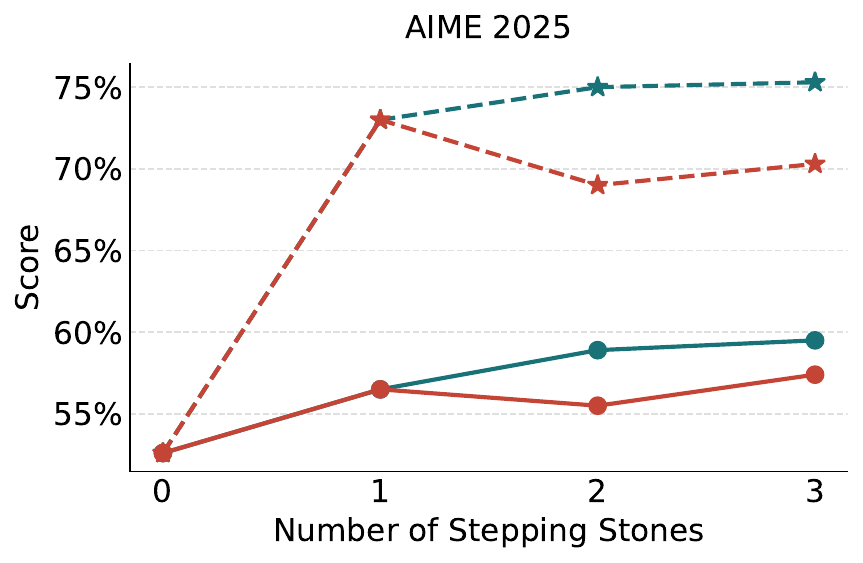}
  \end{minipage}\hfill
  \begin{minipage}[t]{0.32\textwidth}
    \centering
    \includegraphics[width=\linewidth]{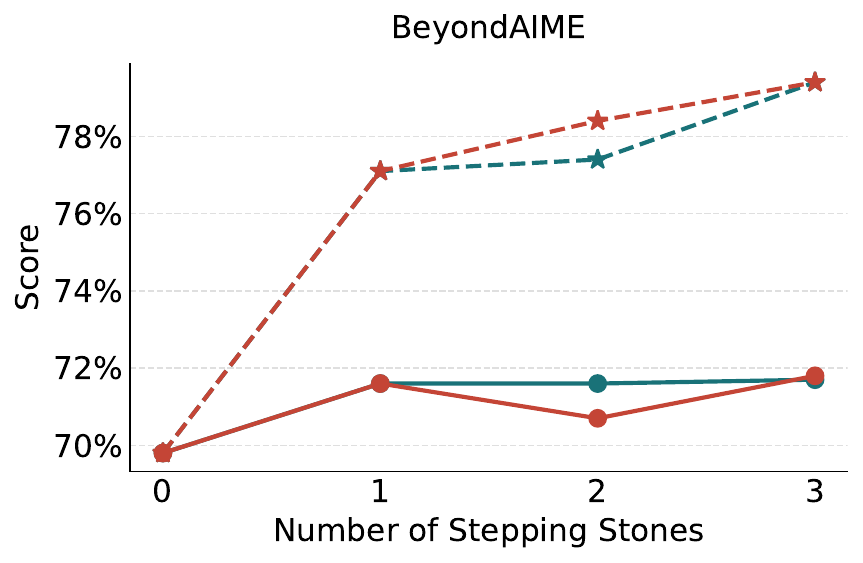}
  \end{minipage}

  \caption{\small Multi-stone ARQ with Qwen3-8B post-trained as a stone generator.}
  \label{fig:multistone-dpo-qwen8b}
\end{subfigure}

\vspace{0.8em}

\begin{subfigure}{\textwidth}
  \centering
  \begin{minipage}[t]{0.32\textwidth}
    \centering
    \includegraphics[width=\linewidth]{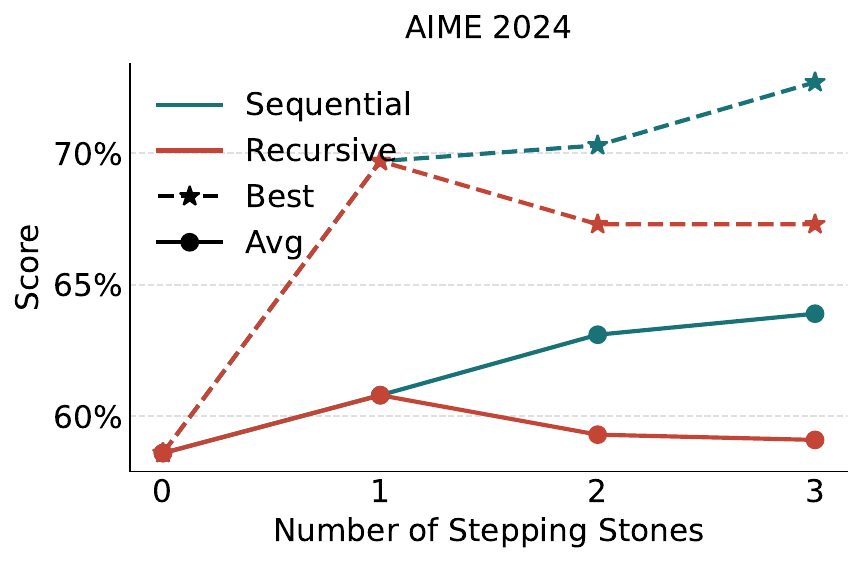}
  \end{minipage}\hfill
  \begin{minipage}[t]{0.32\textwidth}
    \centering
    \includegraphics[width=\linewidth]{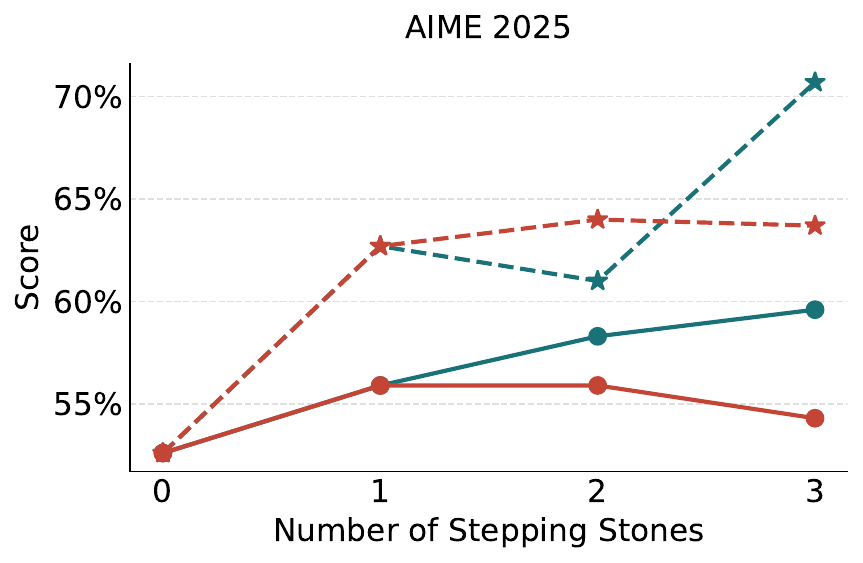}
  \end{minipage}\hfill
  \begin{minipage}[t]{0.32\textwidth}
    \centering
    \includegraphics[width=\linewidth]{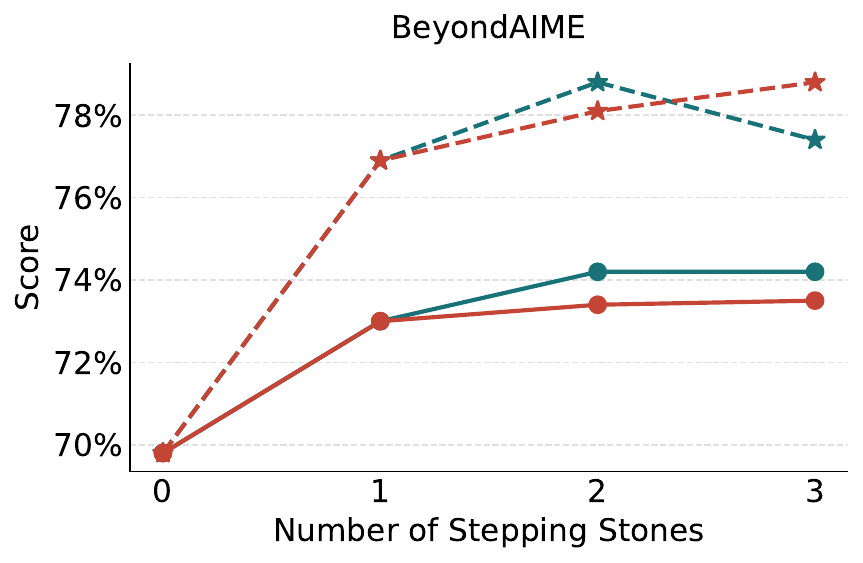}
  \end{minipage}

  \caption{\small Multi-stone ARQ with Qwen2.5-32B post-trained as a stone generator.}
  \label{fig:multistone-dpo-qwen32b}
\end{subfigure}

\caption{Performance of sequential multiple stones and recursive multiple stones on AIME 2024 (left), AIME 2025 (middle), and BeyondAIME (right). \textbf{Best} is conditioned on the best generated stones for the target questions, and \textbf{Avg} is averaged over 20 stones.}
\label{fig:multistone-block}
\end{figure}

For the multiple stone experiments in \cref{sec:more-stones}, we propose two possible paradigms of extending the number of stepping stones, namely sequential stepping stones and recursive stepping stones. We compare these two paradigms using three stone generators and report the average performance over three datasets (AIME 2024, AIME 2025, and BeyondAIME) in \cref{fig:multi-stone}. To compare and analyze the performance on each individual dataset, we plot the performance of GPT-120B, Qwen3-8B post-trained, and Qwen3-32B post-trained in \cref{fig:multistone-dpo-gpt120b}, \cref{fig:multistone-dpo-qwen8b} and \cref{fig:multistone-dpo-qwen32b}, respectively.

\section{The Complete List of Prompts}
\label{sec:appendix-prompt}

In this section we provide the prompts used for our experiments in \cref{sec:good-stone} and \cref{sec:more-stones}.  The prompt to generate a stepping stone in ARQ is shown in \cref{fig:stone-gen-prompt-full}. The prompts to solve the stepping stone and the target problem in ARQ are displayed in \cref{fig:stone-solve-prompt} and \cref{fig:final-solve-prompt} respectively. The prompt to generate a random question for the \emph{Rand} baseline is shown in \cref{fig:stone-random-prompt}. For the experiments involving multiple stones, we use the prompt in \cref{fig:sequential-stone-prompt} to generate stones sequentially and reuse the prompt in \cref{fig:stone-gen-prompt-full} to generate stones recursively by treating the previously generated stone as the new target problem. The prompts in \cref{fig:stone-solve-prompt} and \cref{fig:final-solve-prompt} are used to solve the stepping stones and final problems in the multi-stone experiments. For all prompts that require us to parse out the generated content, such as in stone generation, we instruct the LLMs to wrap the final output in YAML for easy parsing.

\begin{figure}
    \centering
    \begin{tcolorbox}[
        width=1\linewidth,
        colback=gray!7,
        colframe=white!0,
        title=Prompt for Stepping Stone Generation,
        fonttitle=\bfseries,
        boxrule=0.5pt
    ]
    \small
    \textbf{Task:}\\
    We are trying to write some subproblems that may be helpful or inspirational
    for solving the final problem. The subproblems should be stepping stones
    towards solving the final problem.\\
    
    A subproblem can take one or a few of the following roles: \\
    \textbullet\ a simpler version of the final problem; \\
    \textbullet\ a special case of the final problem; \\
    \textbullet\ a practice for some methods that will be useful for solving the final problem. \\    
    \rule{\linewidth}{0.4pt}\\
    \textbf{Subproblem:}\\
    \texttt{\textless Placeholder\textgreater}\\
    \textbf{Final problem:}\\
    \texttt{\{problem\}}\\
    \rule{\linewidth}{0.4pt}\\
    \textbf{Instructions:} \\
    1. Be clear and \textbf{self-contained}, someone seeing only the subproblem should understand what's being asked. \\
    2. You do not need to solve the proposed subproblem. \\
    3. Write the subproblem in the \textbf{same tone and format} as the final problem. \\
    \rule{\linewidth}{0.4pt}\\
    \textbf{Output format:}\begin{verbatim}
```yaml
problem: |
    ...
```
\end{verbatim}
    \end{tcolorbox}
    \caption{Prompt used to generate stepping stone problems in ARQ. The \texttt{\{problem\}} is replaced by the target problem.}
    \label{fig:stone-gen-prompt-full}
\end{figure}

\begin{figure}
    \centering
    \begin{tcolorbox}[
        width=1\linewidth,
        colback=gray!7,
        colframe=white!0,
        title=Prompt for Solvers,
        fonttitle=\bfseries,
        boxrule=0.5pt
    ]
    \small
    \textbf{Question:} \\
    \newline
    \texttt{\{question\}}\\
    \newline
    \rule{\linewidth}{0.4pt}
    Please reason step by step to solve the question above.\\
    For the final solution, include critical details and put your final answer within \texttt{\textbackslash boxed\{\}}.\\
    If the final answer is a fraction, please reduce to the simplest form.\\
    Don't use commas when writing out large numbers.
    \end{tcolorbox}
    \caption{Prompt used to solve stepping stone problems in ARQ. The \texttt{\{question\}} is replaced by the stepping stone problem.}
    \label{fig:stone-solve-prompt}
\end{figure}

\begin{figure}
    \centering
    \begin{tcolorbox}[
        width=1\linewidth,
        colback=gray!7,
        colframe=white!0,
        title=Prompt for Solvers,
        fonttitle=\bfseries,
        boxrule=0.5pt
    ]
    \small
    Study the following example problems and their solutions.\\
    
    \textbf{Example:}\\

    \texttt{\{stone problem\}}\\

    \textbf{Solution to Example:} \\

    \texttt{\{stone solution\}}\\
    
    \rule{\linewidth}{0.4pt}\\
    
    \textbf{Task}: Reason step by step to solve the following final problem, and put your final answer within \texttt{\textbackslash boxed\{\}}.\\
    If the final answer is a fraction, please reduce to the simplest form. \\
    Don't use commas when writing out large numbers.\\
    
    \textbf{Final Problem:} \\
    \newline
    \texttt{\{question\}}
    \end{tcolorbox}
    \caption{Prompt used to solve target problems in ARQ. The \texttt{\{stone problem\}}, \texttt{\{stone solution\}}, and \texttt{\{question\}} are replaced by the stepping stone problem, the solution to the stepping stone problem, and the target problem.}
    \label{fig:final-solve-prompt}
\end{figure}

\begin{figure}[h]
    \centering
    \begin{tcolorbox}[
        width=1\linewidth,
        colback=gray!7,
        colframe=white!0,
        title=Prompt for Solvers,
        fonttitle=\bfseries,
        boxrule=0.5pt
    ]
    \small
    \textbf{Task:}\\
    Please randomly generate an AMC or AIME style math problem to help student practice.\\
    \rule{\linewidth}{0.4pt}\\
    1. Be clear and \textbf{self-contained}.\\
    2. You do not need to solve the proposed problem.\\
    \rule{\linewidth}{0.4pt}\\
    \textbf{Output format:}\begin{verbatim}
```yaml
problem: |
    ...
```
\end{verbatim}
    \end{tcolorbox}
    \caption{Prompt used to generate a random stepping stone in the \emph{Rand} baseline.}
    \label{fig:stone-random-prompt}
\end{figure}

\begin{figure}
    \centering
    \begin{tcolorbox}[
        width=1\linewidth,
        colback=gray!7,
        colframe=white!0,
        title=Prompt for Solvers,
        fonttitle=\bfseries,
        boxrule=0.5pt
    ]
    \small
    \textbf{Task}:
    
    We are trying to write some subproblems that may be helpful or inspirational for solving the final problem. The subproblems should be stepping stones towards solving the final problem.\\
    
    Each subproblem can take one of the following roles: \\
    \textbullet\  a simpler version of the final problem;\\
    \textbullet\  a special case of the final problem;\\
    \textbullet\  a practice for some methods that will be useful for solving the final problem.\\
    
    We have proposed a few subproblems. Please propose a new subproblem that takes a step closer towards solving the problem.\\
    \rule{\linewidth}{0.4pt}\\
    Subproblem \{1\}:\\
    \{subproblem 1\}\\

    Subproblem \{2\}:\\
    \{subproblem 2\}\\

    ...\\
    
    Subproblem \{k\}:\\
    \texttt{<Placeholder>}\\
    
    Final problem:\\
    \{problem\}

    \rule{\linewidth}{0.4pt}\\
    
    \textbf{Instructions:}\\
    
    1. Be clear and \textbf{self-contained}, someone seeing only the subproblem should understand what's being asked.\\
    2. You do not need to solve the proposed subproblem. \\
    3. Write the subproblem in the \textbf{same tone and format} as the final problem. \\
    
    \rule{\linewidth}{0.4pt}\\
    
   \textbf{Output format:}\\
   
    \verb|```| yaml\\
    subproblem: $|$\\
    \quad ...\\
    \verb|```| 
    \end{tcolorbox}
    \caption{Prompt used to generate the $k$-th sequential problem in ARQ.}
    \label{fig:sequential-stone-prompt}
\end{figure}

\section{The Use of Large Language Model}
\label{sec:llm-usage}
Large language model is used in our study as a general-purpose assist tools and we use it for checking grammar mistakes and fixing Latex compile errors.

\section{Dataset Credits and Licenses}
\label{sec:dataset-licenses}
We use the following datasets as test benchmarks. We acknowledge the original dataset creators and list the corresponding sources and licenses below:

\noindent
\textbf{AIME pre-2024}: 
\url{https://huggingface.co/datasets/gneubig/aime-1983-2024}. \\
Licensed under CC0: Public Domain.

\noindent
\textbf{AIME 2024}: 
\url{https://huggingface.co/datasets/gneubig/aime-1983-2024}. \\
Licensed under CC0: Public Domain.

\noindent
\textbf{AIME 2025}: 
\url{https://huggingface.co/datasets/opencompass/AIME2025}. \\
Licensed under the MIT License.

\noindent
\textbf{BeyondAIME}: 
\url{https://huggingface.co/datasets/ByteDance-Seed/BeyondAIME}. \\
Licensed under the MIT License.

\end{document}